\documentclass[10pt,journal,compsoc]{IEEEtran}
\usepackage{etoolbox}
\usepackage{float}
\usepackage{graphicx}
\usepackage{epstopdf}
\usepackage{placeins}
\usepackage{array}
\DeclareGraphicsExtensions{.eps,.pdf} 
\graphicspath{ {figures/} }
\usepackage{bm}
\usepackage{balance}
\usepackage[ruled,vlined]{algorithm2e}
\usepackage{amssymb,amsmath,mathtools}
\usepackage{amsfonts}
\usepackage{ragged2e}
\usepackage{makecell}
\usepackage{ragged2e}
\usepackage{booktabs,tabularx}
\usepackage{pdflscape}  
\newcolumntype{Y}{>{\raggedright\arraybackslash}X}
\renewcommand{\arraystretch}{1.08}

\newcommand{\InlineCenteredAbstract}[2]{%
  \\[1.4\baselineskip]
  \begin{center}
    \begin{minipage}{0.95\textwidth}
      \small
      \justifying
      \textbf{Abstract—} #1\par
      \vspace{0.5\baselineskip}
      \textbf{Index Terms—} {Electrocardiography (ECG), Masked Vision Transformer, Self-Supervised Learning, Foundation Model.}
    \end{minipage}
  \end{center}%
  \vspace{-1\baselineskip}
}

\usepackage{algorithmicx}
\usepackage{color}
\usepackage{multirow}
\usepackage[caption=false,font=footnotesize]{subfig}
\usepackage[nocompress]{cite}

\usepackage{hyperref}
\usepackage{xr-hyper}  
\makeatletter
\newcommand\restartchapters{\par
  \setcounter{chapter}{0}%
  \setcounter{section}{0}%
  \gdef\@chapapp{\chaptername}%
  \gdef\thechapter{\@arabic\c@chapter}}
\makeatother

\usepackage[utf8]{inputenc}
\usepackage[T1]{fontenc}
\usepackage{amsthm}

\usepackage{wrapfig}
\makeatletter
\g@addto@macro\normalsize{%
 \setlength\abovedisplayskip{4pt}
 \setlength\belowdisplayskip{4pt}
 \setlength\abovedisplayshortskip{4pt}
 \setlength\belowdisplayshortskip{4pt}
}
\makeatother
\makeatletter
\def\endthebibliography{%
	\def\@noitemerr{\@latex@warning{Empty `thebibliography' environment}}%
	\endlist
}
\makeatother

\newlength{\BioGap}
\newcommand{\biogap}{\vspace{\BioGap}}

\title{ECG-Soup: Harnessing Multi-Layer Synergy for ECG Foundation Models}

\author{Phu~X.~Nguyen,
        Huy~Phan,
        Hieu~Pham,
        Christos~Chatzichristos,
        Bert~Vandenberk,
        and Maarten~De~Vos
\IEEEcompsocitemizethanks{
\IEEEcompsocthanksitem Phu X. Nguyen is with STADIUS Center for Dynamical Systems, Signal Processing and Data Analytics, the Department of Electrical Engineering (ESAT), KU Leuven, Leuven 3001, Belgium. Email: \textit{phu.nguyen@kuleuven.be}
\IEEEcompsocthanksitem Huy Phan is with Meta Reality Labs, Paris 75002, France.
\IEEEcompsocthanksitem Hieu Pham is with VinUni--Illinois Smart Health Center, VinUniversity, Hanoi, Vietnam.
\IEEEcompsocthanksitem Christos Chatzichristos is with STADIUS Center for Dynamical Systems, Signal Processing and Data Analytics, the Department of Electrical Engineering (ESAT), KU Leuven, Leuven 3001, Belgium.
\IEEEcompsocthanksitem Bert Vandenberk is with the Department of Cardiovascular Sciences, KU Leuven and with the Department of Cardiology, University Hospitals Leuven, Leuven 3001, Belgium.
\IEEEcompsocthanksitem Maarten De Vos is with STADIUS Center for Dynamical Systems, Signal Processing and Data Analytics, the Department of Electrical Engineering (ESAT) and with the Department of Development \& Regeneration, KU Leuven, Leuven 3001, Belgium.
\protect\\
}
\InlineCenteredAbstract{%
Transformer-based foundation models for Electrocardiograms (ECGs) have recently achieved impressive performance in many downstream applications. However, the internal representations of such models across layers have not been fully understood and exploited. An important question arises: Does the final layer of the pretrained Transformer model, the \emph{de facto} representational layer, provide optimal performance for downstream tasks? Although our answer based on empirical and theoretical analyses for this question is negative, we propose a novel approach to leverage the representation diversity of the model's layers effectively. Specifically, we introduce a novel architecture called Post-pretraining Mixture-of-layers Aggregation (PMA), which enables a flexible combination of the layer-wise representations from the layer stack of a Transformer-based foundation model. We first pretrain the model from ECG signals using the 1-dimensional Vision Transformer (ViT) via masked modeling. In downstream applications, instead of relying solely on the last layer of the model, we employ a gating network to selectively fuse the representations from the pretrained model's layers, thereby enhancing representation power and improving performance of the downstream applications. In addition, we extend the proposed method to the pretraining stage by aggregating all representations through group-wise averaging before feeding them into the decoder-based Transformer. Extensive experimental results demonstrate that our proposed models outperform other self-supervised learning (SSL) baselines on various arrhythmia classification benchmarks with different settings (i.e., in-distribution and out-of-distribution datasets). The proposed approaches obtain a macro AUC exceeding 94\% for 71 ECG conditions and show strong generalization in various application settings. Furthermore, our pretrained model was utilized as the backbone and ranked 1st in the 2025 George B. Moody PhysioNet Challenge on Detection of Chagas Disease from ECG over 630 participants from 111 teams, demonstrating its strong real-world performance. Finally, the detailed analysis further consolidates and underscores the crucial role of the multi-layer representation mixture.
}

\thanks{The work does not relate to H.~Phan's position at Meta.}
\thanks{Source code is available here: \url{https://github.com/Xuanphu108/ecg_ssl}}
}

\begin{document}
\justifying
\maketitle
\section{INTRODUCTION} \label{Introduction}
Cardiovascular disease (CVD) is the leading cause of death globally, accounting for 32\% of all deaths according to The World Health Organization (WHO) statistics in 2019 \cite{anbalagan2023}. With its non-invasive nature and ability to reflect the heart's electrical activity, the electrocardiogram is a key diagnostic tool in clinical practice \cite{marko2017, saman2018}. However, traditional ECG analysis is mainly based on human experts prone to errors and delays. Deep learning models with a supervised learning paradigm \cite{awni2019, antonio2020, zachi2019, nils2021, junaid2022, muhammad2023, shunxiang2023, wei2024, muhammad2024, yang2023, hany2024, xiaoya2024, xiaoyu2021} have shown effectiveness in automating ECG analysis and aiding CVD diagnosis. The supervised learning paradigm, however, faces inherent limitations such as reliance on large-scale annotated data, lack of generalization ability, and susceptibility to data heterogeneity. To address these issues, self-supervised learning (SSL) methods have been proposed. These methods leverage unlabeled ECG data to train robust foundational models, which are then fine-tuned for specific downstream tasks. This approach promises improved generalization and reduced reliance on manually annotated data. 

Self-supervised ECG learning (eSSL) generally includes two primary methods: contrastive learning \cite{bryan2021, dani2021, sahar2022, crystal2022, pritam2022, duc2023, ning2024} and generative learning \cite{huaicheng2022, wenrui2024, ya2024, kuba2024, sehun2024, yeongyeon2024}. The former learns by pulling together similar pattern representations and pushing apart different pattern representations. It is often based on data augmentation techniques, as a result, frequently distorts the semantic meaning of the original ECG signal \cite{xiang2024, yeongyeon2024, che2024}. In contrast, the latter learns by reproducing the original signal, thereby retaining more semantic information. It is better at preserving the semantic information in ECG data, as it learns data representation by reconstructing all or part of the original input. However, the generative approach also has its own issues. By reconstruction-based learning, it often overlooks high-level semantics which are crucial for downstream tasks. This can lead to suboptimal performance when using generative pretrained models for classification \cite{yeongyeon2024, che2024, he2022, jun2023}.

Contrastive predictive coding of the ECG (CPC) is explored in \cite{temesgen2022}. Unlike the contrastive approaches above, CPC predicts multiple future time steps using powerful autoregressive models in the latent space \cite{aaron2019}. The goal is to force the model to focus on the global structure and disregard low-level information. Furthermore, the CPC method does not require complicated augmentations, thus preserving the semantic information of the ECG data. Despite these advantages, this method heavily relies on negative samples (i.e., unrelated ECG samples not matching the true future). The model may fail to learn meaningful representations if negative samples lack diversity. Additionally, CPC typically employs autoregressive models (e.g., recurrent neural networks) for future predictions, making capturing long-range dependencies challenging. To learn global information and preserve the semantic meaning of ECG signals, Transformer-based architectures have been utilized in \cite{wenrui2024, ya2024, sehun2024, kuba2024, yeongyeon2024}, thanks to their attention mechanism. 

Recently, SSL-based vision transformer (ViT) models have been increasingly used for ECG foundation models. In this line of work, the representation obtained from the last layer of a pretrained model has been the default for downstream tasks. No studies have examined the intermediate layers' representation power for downstream tasks. We show that the layers of a pretrained ViT model often exhibit diverse distributions, and there is no guarantee that the last layer will provide the best representation of the downstream tasks. Our analyses also indicate that the representation power is lowest in the first layers, increases and peaks in the middle layers, and then decreases slightly towards the last layers. Motivated by this, we explore ViT's intermediate layers as alternatives to the last layer for downstream tasks. We then propose methods, both in-pretraining and post-pretraining, to dynamically aggregate the representations across different layers of a ViT model to produce the collective representation. Our contributions are as follows.
\begin{itemize}
    \item Through empirical and theoretical analyses, we illuminate the representation power of intermediate layers of a pretrained ViT model for ECG downstream tasks. 
    \item We propose post-pretraining aggregation methods based on (i) pooling and (ii) a Mixture of Layers model to fuse the representations from different layers of a pretrained ViT model for ECG downstream tasks.
    \item In the pretraining task, we further investigate the impact of aggregating intermediate representations from different layers in the ViT model encoder before feeding them to the Transformer decoder. 
    \item Through extensive experiments, we show that (i) our proposed models outperform SSL baselines on various downstream arrhythmia classification benchmarks with different settings (i.e., in-distribution vs. out-of-distribution (OOD), linear probing vs. fine-tuning) and that (ii) the learned representations are meaningful for ECG signals. In addition to extensive evaluations on multiple datasets, our pretrained model achieved the 1st place in the 2025 George B. Moody PhysioNet Challenge (Detection of Chagas Disease from ECG), highlighting its robustness and generalization capability in a real-world benchmark setting \cite{matthew2025}.  
\end{itemize}

\section{The Backbone Model}
\subsection{1-Dimensional Vision Transformer (ViT1d)}
ECG signals are usually long time series. Directly applying the conventional Transformer model at each time point significantly increases the computational cost due to its self-attention mechanism and limits the ability to exploit important morphological features of the signal. ECG components, such as the P, T, and U waves, contain clinically meaningful shape information that can be overlooked if the model processes the signal only at the time point level rather than in contextual segments. In this paper, we employed as the backbone network an 1-dimensional ViT for fixed-length 12-lead ECG signals \cite{yeongyeon2024}. 

\subsection{Masked Vision Transformer}
We applied a Spatio-Temporal Masked Modeling (STMEM) strategy to pretrain the ECG foundation model \cite{yeongyeon2024}. Specifically, each 10-second ECG signal (12 leads, 100 Hz) was divided into patches and mapped into the input embedding sequence for the ViT architecture. In the pretraining phase, 75\% of the patches were masked, and only the remaining patches were passed through the encoder block to learn the feature representation. The encoder consists of twelve stacked Transformer layers, in which the unmasked embeddings were added to the learned lead embeddings, then passed through self-attention layers to generate the global context representation. The decoder took the output representation from the encoder, projected it into a smaller feature space, and reconstructed the masked patches with a lightweight decoder Transformer (i.e., four stacked Transformer layers) for each lead, to prevent the model from “leaking” information between leads. The training objective was to minimize the reconstruction error (MSE) between the original ECG signal and the reconstructed patches. The implementation details of the backbone model, including the encoder-decoder architecture and training objectives, are provided in the Appendix \ref{Appendix:A}.

This STMEM-based pretrained ViT backbone served as the foundation for our subsequent analysis. To better understand how information is organized across layers and to motivate the design of our proposed multi-layer representation aggregation framework, we next investigate the representational properties of hidden layers in the pretrained model.

\section{Effect of hidden layers of pretrained ViTs}

\begin{figure}[H]
  \centering
  \subfloat[Layer-wise macro AUC]{%
    \includegraphics[width=0.81\linewidth]{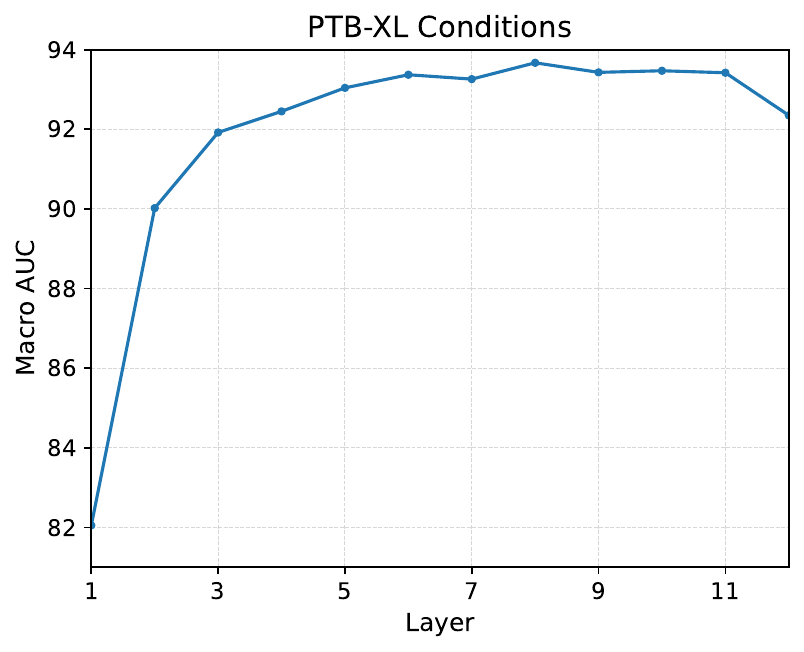}%
    \label{macro_auc_single}
  }
  \par\medskip
  \subfloat[Average cosine similarity through inner layers]{%
    \includegraphics[width=0.81\linewidth]{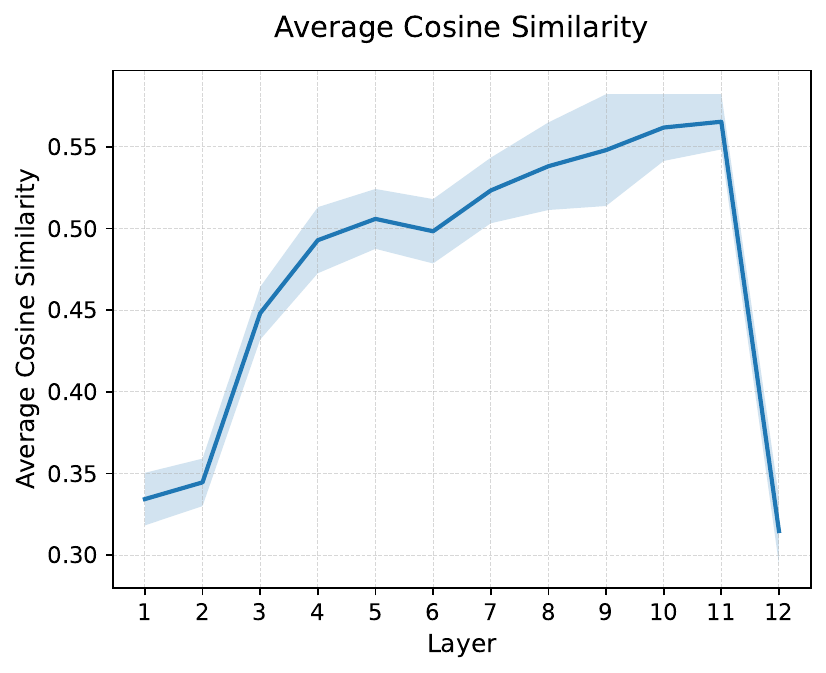}%
    \label{single_avg_cosine}
  }

  \caption{Representation analysis across layers of STMEM-based pretrained ViT on the PTB-XL dataset, a large publicly available clinical 12-lead ECG database.}
  \label{representation}
\end{figure}

To investigate the representational power of the intermediate layers compared to the final layer in the pretrained ViT, we extracted features from different layers and then used linear probing to evaluate them on downstream tasks. This approach allows us to observe the variation in representational quality with network depth directly, thereby elucidating the prominent role of intermediate layers in capturing ECG information.

The results presented in Figure \ref{macro_auc_single} show that the representations at the final layers of the model do not provide optimal performance for the downstream classification tasks. The performance improves gradually from the early layers, peaks at the middle layers, and degrades at deeper layers. This is mainly because the representations at early layers are still raw and discrete, reflecting the uncertainty of information \cite{maithra2021, shentong2023, jun2019, oscar2025}. In the middle layers, the models begin to accumulate and aggregate information in depth, allowing it to learn the hidden relationships between different components of the signal, such as the morphological correlation between the ECG waves (P, QRS, T) and their associated time intervals, thus creating highly generalizable representations that are more suitable for downstream tasks. At last layers, the mutual information between patches is degraded as the model shifts its optimal goal to reconstruct the original signal, thereby reducing the value of the representation for the classification task. 

\section{Proposed method}
Linear probing analysis shows that the layers within the pretrained ViT model contribute unevenly to downstream classification performance: performance typically increases from the early layer, peaks in the middle layer, and then declines in the last layer. Such a phenomenon is related to the change in the correlation level between patches across layers, quantified by the average cosine similarity between tokens at each layer \cite{chengyue2021}, as shown in Figure \ref{single_avg_cosine}: \textit{"The average cosine similarity increases gradually from the early layers, converges at the middle layers, and decreases at the last layers"}. This also means that the representation of the middle layer is often more informative than that of other layers. To consolidate this observation, we conducted a mathematical analysis to explain the information transformation process through the layers of the Transformer model trained by the masked modeling mechanism.
\subsection{Theoretical analysis}
We denote:

\begin{itemize}   
    \item $\textbf{x} = [\textbf{x}_1, \textbf{x}_2, \ldots, \textbf{x}_N]$: ECG patch embeddings.

    \item $N$: the number of ECG patches.

    \item $d$: the embedding dimension.

    \item $\mathcal{M}$: the set of masked patches.

    \item $\textbf{H}^{(l)} \in \mathbb{R}^{N \times d}$: the representation matrix at $l$-th layer, where each row $\textbf{h}_i^{\rm T}$ is representation vector of $i$-th patch embedding.

    \item Attention head: $\textbf{Q} = \textbf{H}^{(l)}\textbf{W}_{\rm Q}$, $\textbf{K} = \textbf{H}^{(l)}\textbf{W}_{\rm K}$, $\textbf{V} = \textbf{H}^{(l)}\textbf{W}_{\rm V}$, where $\textbf{W}_{\rm Q}, \textbf{W}_{\rm K}, \textbf{W}_{\rm V} \in \mathbb{R}^{d \times d_{k}}$ are learnable matrices.

    \item Attention matrix: 
    \[
    \textbf{A} = \mathrm{softmax}\left(\frac{\textbf{QK}^\top}{\sqrt{d_k}}\right) \in \mathbb{R}^{N \times N},
    \]
    where 
    \[
    a_{ij} = 
    \frac{\exp\left(\textbf{q}_i^\top \textbf{k}_j / \sqrt{d_k}\right)}
    {\sum_{t=1}^N \exp\left(\textbf{q}_i^\top \textbf{k}_t / \sqrt{d_k}\right)},
    \]
    each $i$-th row of the attention matrix is a stochastic distribution on $\{1; 2; ...; N\}$: $a_{ij} \geq 0 \quad$ and $\quad \sum_j a_{ij} = 1$.
\end{itemize}

It is important to note that the analysis concerns the effect of self-attention on the correlation between patch embeddings at each layer in ViT, while the matrix $\textbf{W}_{V}$ is only used to project the feature dimension. Therefore, we may ignore $\textbf{W}_{\rm V}$ or consider $\textbf{V} = \textbf{H}^{(l)}\textbf{W}_{\rm V}$ and then investigate the self-attention mechanism separately.

To analyze the effect of the self-attention mechanism, we consider the overlap information between patches in each layer, which is given by 

\begin{equation} \label{eq:1}
\begin{aligned}
\Delta(\textbf{H}) = \max_{x,y} \| \textbf{h}_x - \textbf{h}_y \|,
\end{aligned}
\end{equation}
where $\|.\|$ is a norm on $\mathbb{R}^{d}$.

\noindent The Dobrushin contraction coefficient as

\begin{equation} \label{eq:2}
\begin{aligned}
\delta(\textbf{A}) = 1 - \min_{i,j} \sum_{k=1}^N \min \{ a_{ik}, a_{jk} \},
\end{aligned}
\end{equation}
where $\sum_{k=1}^N \min \{ a_{ik}, a_{jk} \}$ is the overlap between two distributions. It is easy to recognize some properties:

\begin{itemize}   
    \item If two distributions are very similar, the overlap is close to 1, and consequently $\delta(\textbf{A})$ becomes small.

    \item If two distributions are very different, $\delta(\textbf{A})$ becomes large.

    \item Since $\textbf{a}_i$ and $\textbf{a}_j$ are two distributions, $0 \leq \sum_{k=1}^N \min \{ a_{ik}, a_{jk} \} \leq 1$. This leads to $0 \leq \delta(\textbf{A}) \leq 1$. 
\end{itemize}

\noindent \textbf{Lemma 1:} For any vector $\textbf{b}_{k} \subset \mathbb{R}^d$ and row $i, j$; the following inequality holds:

\begin{equation} \label{eq:3}
\begin{aligned}
\left\| \sum_{k} a_{ik} \textbf{b}_k - \sum_{k} a_{jk} \textbf{b}_k \right\|
\leq \delta(\textbf{A}) \max_{x,y} \| \textbf{b}_x - \textbf{b}_y \|.
\end{aligned}
\end{equation}

\textit{Proof:} See Appendix \ref{Appendix:B}.

\noindent Applying for $\textbf{b}_k = \textbf{v}_k$ (or $\textbf{b}_k = \textbf{h}_k$ if ignore $\textbf{W}_{\rm V}$): 

\begin{equation} \label{eq:4}
\begin{aligned}
\Delta(\textbf{H}^{(l+1)}) \leq \delta(\textbf{A})\,\Delta(\textbf{H}^{(l)}).
\end{aligned}
\end{equation}

\noindent Repeating $L$ times:

\begin{equation} \label{eq:5}
\begin{aligned}
\Delta(\textbf{H}^{(L)}) 
\leq \left( \prod_{l=0}^{L-1} \delta(\textbf{A}^{(l)}) \right) \Delta(\textbf{H}^{0}). 
\end{aligned}
\end{equation}

\noindent Since $\delta(\textbf{A}) \leq 1$, $\Delta(\textbf{H}^{(L)})$ decays exponentially. This also means that the correlation between ECG patch embeddings increases through layers. However, since the pretrained ViT model is optimized using the mask modeling with the MSE loss function, the correlation between embedding patches in the last layers tends to decrease. The reason is that MSE encourages the model to reproduce the original signal accurately, thereby forcing the embedding patches to be more clearly separated in the feature space to carry more independent information. 

\begin{equation} \label{eq:6}
\begin{aligned}
\mathcal{L}_{\text{MSE}} = \frac{1}{|\mathcal{M}|} \sum_{i \in \mathcal{M}} \left\| \textit{decoder}(\textbf{e}_i, \textbf{h}^{(L)}) - \textbf{x}_i \right\|^2,
\end{aligned}
\end{equation}
where $\textbf{e}_i$ are learnable embeddings of masked patches.

\subsection{Cross-layer aggregation schemes}

Since each layer learns different levels of abstraction, from low-level features in the top layer, high-level semantic information in the middle layer, to sophisticated patterns in the deep layer, multilayer fusion can improve representation quality, reduce overfitting, and improve generalization on ECG data from multiple sources \cite{yuan2023, chenghao2023, martina2023, jinsu2023}. Based on this observation, we proposed three multilayer extraction mechanisms:

\textit{Scheme I - Post-pretraining Pooling-based Aggregation (PPA):}
After pretraining, we take the output embedding from all layers of ViT encoders, perform average grouping, and feed this aggregated feature into the classifier layer.

\begin{figure}[H]
    \centering
    \includegraphics[width=0.49\textwidth]{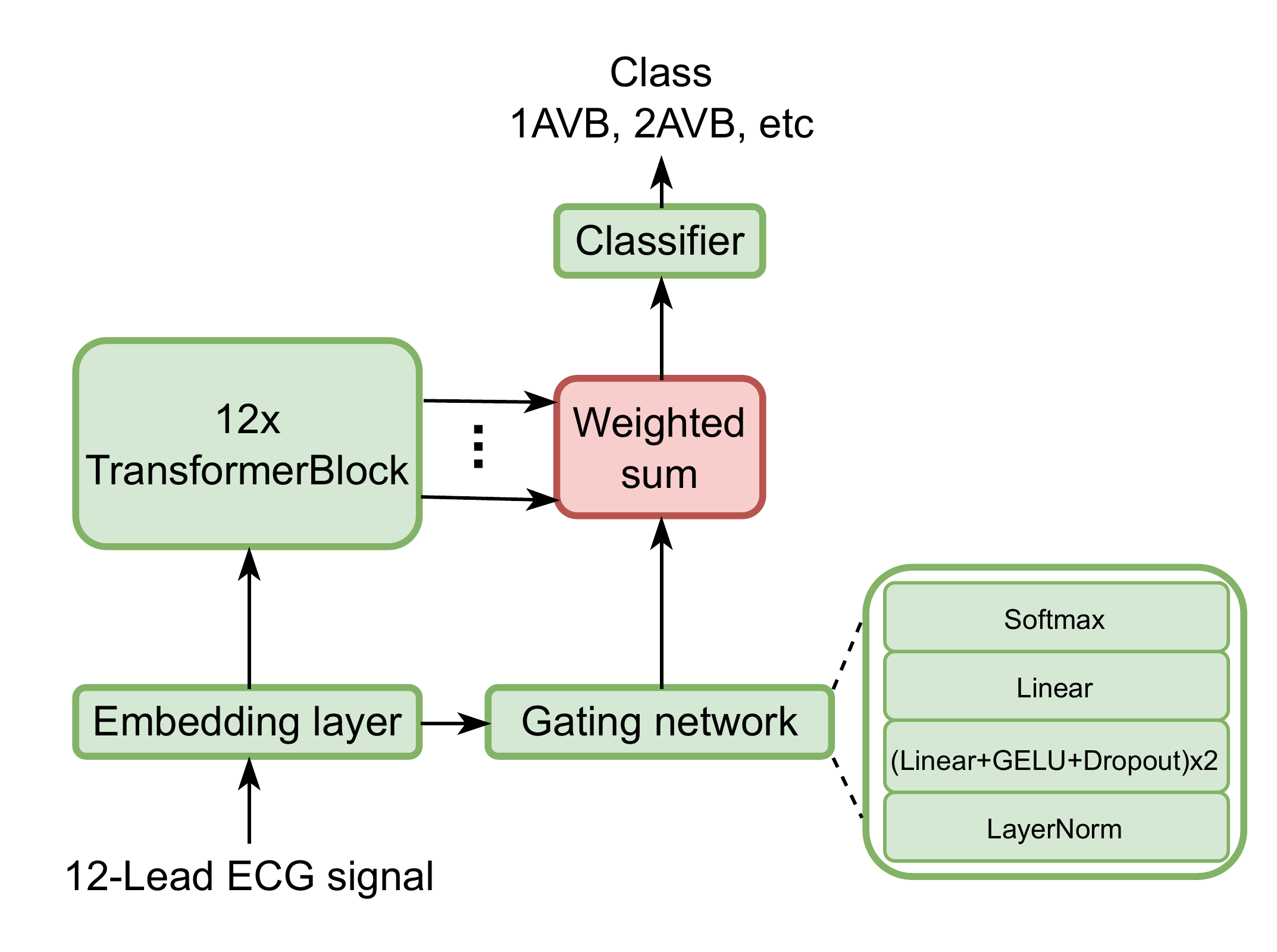}
    \caption{Overview of post-pretraining mixture-of-layers aggregation of pretrained ViT's different layer-wise representations.}
    \label{fig:mol}
\end{figure}

\textit{Scheme II - Post-pretraining Mixture-of-layers Aggregation (PMA):}
Instead of assigning fixed weights, inspired by the Mixture of Experts (MoE) architecture \cite{zixiang2022, weilin2024}, PMA learns a small gating network to calculate soft weights for each layer. The layer embeddings are linearly combined according to the learned weights, automatically allowing the model to select functional layers for each ECG sample. The detailed architecture is illustrated in Figure \ref{fig:mol}.

\textit{Scheme III - In-pretraining Pooling-based Aggregation STMEM (IPASTMEM):}
Unlike PPA/PMA, which is only applied after pretraining, IPASTMEM integrates the pooling mechanism right in the pretraining process: intermediate layers are pooled before being passed to the decoder, which helps spread the gradient more evenly between layers and improves generalization ability, especially in OOD settings.

\subsection{Downstream classification}
In the downstream classification task, we retained the encoder of the masked ViT models (e.g., STMEM and IPASTMEM) and then removed the $[SEP]$ tokens before feeding the representation vector into a classification head. Finally, we added a simple classification head with basic layers, such as batch normalization, dropout, ReLU, and a linear layer to map the output representation to the target labels. The model was applied to the multilabel classification problem on different datasets to identify ECG conditions and heart rhythm types. 

The model's input data was a 12-lead ECG signal with a duration of 10 seconds. This signal was first fed into the encoder to extract the representation vector. Then, this vector was fed into the classifier to predict cardiovascular conditions, $\{\hat{o}_{1}, \ldots, \hat{o}_{C} \}$. The loss function used during training is defined as follows:
\begin{align}
\label{eq:17}
\mathcal{L}_{\text{CE}} = \frac{1}{C} \sum_{c=1}^{C} \left( -o_c \log(\hat{o}_c) \right), 
\end{align}
where ${\bf o} = \{o_{1}, \ldots, o_{C} \}$ denotes the actual multi-label target and $C$ is the number of labels.

\section{Experimental Settings}
In this section, we describe in detail the experimental settings in this paper, including the implementation method, the baseline models, and the datasets used for our experiments.

\subsection{Datasets}

\begin{table}[H]
\centering
\caption{Summary of the datasets.}
\label{tab:1}
\begin{tabular}{|l|c|c|c|}
\hline
\textbf{Dataset} & \textbf{Samples} & \textbf{Sample rate} & \textbf{Duration}\\
\hline
\textbf{Pretraining: \textit{All}} & 400,365 & multiple & multiple \\
- \textit{CinC2020} \cite{erick2021} & 43,093 & multiple & multiple \\
- \textit{Chapman} \cite{jianwei2020} & 10,646 & 500 Hz & 10s \\
- \textit{Ribeiro-test} \cite{antonio2020} & 827 & 400 Hz & 7s, 10s \\
- \textit{CODE-15} \cite{ribeiro2021} & 345,799 & 400 Hz & 7s, 10s\\
\hline
\textbf{Evaluation:} & & & \\
- \textit{PTB-XL} \cite{patrick2020} & 21,837 & 100 Hz, 500 Hz & 10s \\
- \textit{Chapman} \cite{jianwei2020} & 10,646 & 500 Hz & 10s \\
\hline
\end{tabular}
\end{table}

This paper utilized five 12-lead ECG datasets collected from various countries and demographics for pretraining and downstream tasks. To ensure consistency throughout the training and evaluation process, all ECG signals were normalized to the same input format with a fixed sampling rate of 100 Hz and a signal length of 10 seconds. Specifically, recordings with durations exceeding 10 seconds were truncated. Conversely, shorter signals were zero-padded to ensure a standard length, facilitating consistent model building and training across datasets. In the pretraining phase, the model was trained on a combined set of four datasets, namely PhysioNet / Computing in Cardiology Challenge 2020 (CinC2020) \cite{erick2021}, Chapman (Zheng) \cite{jianwei2020}, CODE-test (Ribeiro2020) \cite{antonio2020}, and CODE-15 \cite{ribeiro2021}, with a total of 400,365 records. For the downstream phase, the models were fine-tuned in two datasets, PTB-XL \cite{patrick2020} and Chapman, to evaluate the fitness and generalizability of the learned representations on different datasets and settings. It is essential to note that the CinC2020 dataset was compiled from five different data sources, including PTB-XL. We set up two separate training scenarios on the downstream classification task to evaluate the generalization of pretraining models: in-distribution and OOD on the PTB-XL dataset. Specifically, in the in-distribution scenario, the original version of CinC2020 was kept intact, including data from PTB-XL, and used to train the pretrained models. In contrast, in the OOD scenario, the PTB-XL dataset was removed from CinC2020 before training, and PTB-XL is only used in the downstream phase. This experimental design enables us to evaluate the model's dependence on the training data and its generalization ability when presented with an unseen data distribution during the pretraining phase. All datasets are summarized in Table \ref{tab:1}.

\subsection{Training and evaluation}
In the pretraining phase, the models were trained completely unsupervised (i.e., without using any labels) to learn informative feature representations from ECG data. After completing the pretraining process, the pretrained models were fine-tuned and evaluated on two labeled datasets, PTB-XL and Chapman, representing prosperous clinical data sources. In this phase, we built two classification scenarios that reflect common application goals in practice: (1) classifying all ECG conditions and (2) classifying ECG rhythms, illustrated in Table \ref{tab:2}. Each dataset was partitioned into 10 non-overlapping folds (8 for training, 1 for validation, and 1 for testing) \cite{temesgen2022}. To ensure the reliability and reproducibility of the results, we controlled for randomness by repeating all experiments with 10 different random seeds across all datasets and classification scenarios. The reported scores were the averages over these runs, reflecting the stability and robustness of the models against variations in weight initialization. Model performance was assessed using both macro- and sample-level metrics, including macro/sample AUC, instance/sample accuracy, and macro/sample F1-score, providing a comprehensive evaluation of the models in multi-label classification settings. 

\subsection{Implementation Details}
Evaluating the effectiveness of our proposed self-supervised learning model requires a comprehensive benchmarking system with various baseline methods. In this study, we established a set of baseline models representing three major approaches in machine learning for electrocardiogram (ECG) signals, including supervised learning from scratch, contrastive learning, and generative learning. A complete list of hyperparameters (batch size, learning rate, warmup steps, etc.) is provided in Table \ref{tab:3}.

\begin{table}[H]
\centering
\caption{ECG Conditions and Rhythms in PTB-XL and Chapman Datasets.}
\label{tab:2}
\begin{tabular}{|l|c|c|}
\hline
\textbf{Dataset} & \textbf{ECG Conditions} & \textbf{ECG Rhythms}\\
\hline
PTB-XL & 71 & 12 \\
\hline
Chapman & 67 & 11 \\
\hline
\end{tabular}
\end{table}

In the group of supervised learning models, we implemented and evaluated three popular baseline architectures: the 1-dimensional XResnet50 (XResnet1d50), a hybrid architecture consisting of 4FC, LSTM, and 2FC (4FC+2LSTM+2FC), as introduced in \cite{nils2021, temesgen2022}, and ViT1d \cite{alexey2021, yeongyeon2024}. These models were trained from scratch on standardized ECG datasets (e.g., PTB-XL condition, PTB-XL rhythm, Chapman condition, and Chapman rhythm), without any pretraining knowledge, to reflect their ability to learn representations from raw data without the advantages of weight initialization. In particular, the XResnet model was selected due to its efficiency in processing one-dimensional signals. At the same time, ViT represented a state-of-the-art Transformer architecture in the field of vision, and 4FC+2LSTM+2FC demonstrated the ability to exploit temporal information to predict the following context. For XResnet1d50 and 4FC+2LSTM+2FC, we used the AdamW optimizer with a fixed learning rate of 0.001 and a weight decay factor set to 0.001. The training process was performed with a binary cross-entropy loss function for multi-label classification, using a constant learning rate and a batch size of 128. For the ViT1d model, training was performed according to the configuration described in Table \ref{tab:3}, but we used a smaller batch size, namely 16.

\begin{table}[H]
  \centering
  \caption{Hyperparameter settings.}
  \label{tab:3}
  \renewcommand{\arraystretch}{1.2}
  \setlength{\tabcolsep}{3.5pt}

  \begin{tabular}{|l|c|c|c|}
    \hline
    \textbf{Hyperparameter} & \textbf{Pretrain} & \textbf{Linear} & \textbf{Fine-tune} \\
    \hline
    Backbone       & ViT     & ViT     & ViT     \\
    Learning rate  & 0.0006  & 0.001   & 0.001   \\
    Batch size     & 128     & 64      & 64      \\
    Epochs         & 800     & 100     & 100     \\
    Optimizer      & AdamW   & AdamW   & AdamW   \\
    LR scheduler   & Cosine  & --      & Cosine  \\
    Warmup steps   & 40      & --      & 5       \\
    \hline
  \end{tabular}
\end{table}

For the group of contrastive learning models, we examined typical methods such as Simple Contrastive Learning (SimCLR) \cite{ting2020} and CPC in the pretraining setting. These models were trained using popular loss functions in contrastive learning, such as noise contrast estimation (NCE) and InfoNCE loss, which aimed to maximize the similarity between the representations of positive pairs while distinguishing them from negative pairs. The optimization process used the AdamW optimizer algorithm. The models were then fine-tuned on multi-label classification tasks to evaluate the generalizability of the learned representation. The implementation details in the pretraining phase were similar to the methods presented in \cite{temesgen2022}. However, in the fine-tuning phase, we used as input an ECG signal of 10 seconds duration, instead of 2.5 seconds as in the fine-tuned models in \cite{temesgen2022}. 

In particular, within the generative learning group, we implemented and evaluated the STMEM, which also served as the foundation for our proposed self-supervised model. In generative learning, the STMEM and our proposed methods used the ViT backbone, where the ECG signal was partitioned into non-overlapping segments (patches) with a patch size of 50 signal points each. During pretraining, 75\% of the patches are masked randomly, and the model was trained to reconstruct the masked patches by minimizing the MSE loss function between the original and reconstructed patches. The implementation details are given in the Table \ref{tab:3}.

\section{Representation transformation in pretrained ViTs}
The previous section provided a preliminary analysis focusing on the STMEM-based pretrained ViT. To develop a more comprehensive and rigorous understanding, this section extends the investigation to multiple pretrained models and diverse evaluation metrics to elucidate the internal mechanisms of representation transformation in pretrained ViTs and their influence on downstream task performance.

\vspace{-1em}
\subsection{Linear probing evaluation across layer:} 
The observation in Figure \ref{representation} is consistent across multiple evaluation metrics, datasets, and pretrained models (i.e., STMEM and IPASTMEM) as shown in Appendix \ref{Appendix:C}, indicating the generality of the information transform across layers trending.

\subsection{Correlation between patch embeddings:}

We used average cosine similarity to measure the correlation between patch embeddings at each layer in pretrained ViT. This metric reflects the structure of the ECG signal representation because the ECG components inherently have close physiological dependencies (such as the relationship between P-QRS-T).  

\begin{figure}[H]
    \centering
    \includegraphics[width=8.6cm,height=4.4cm]{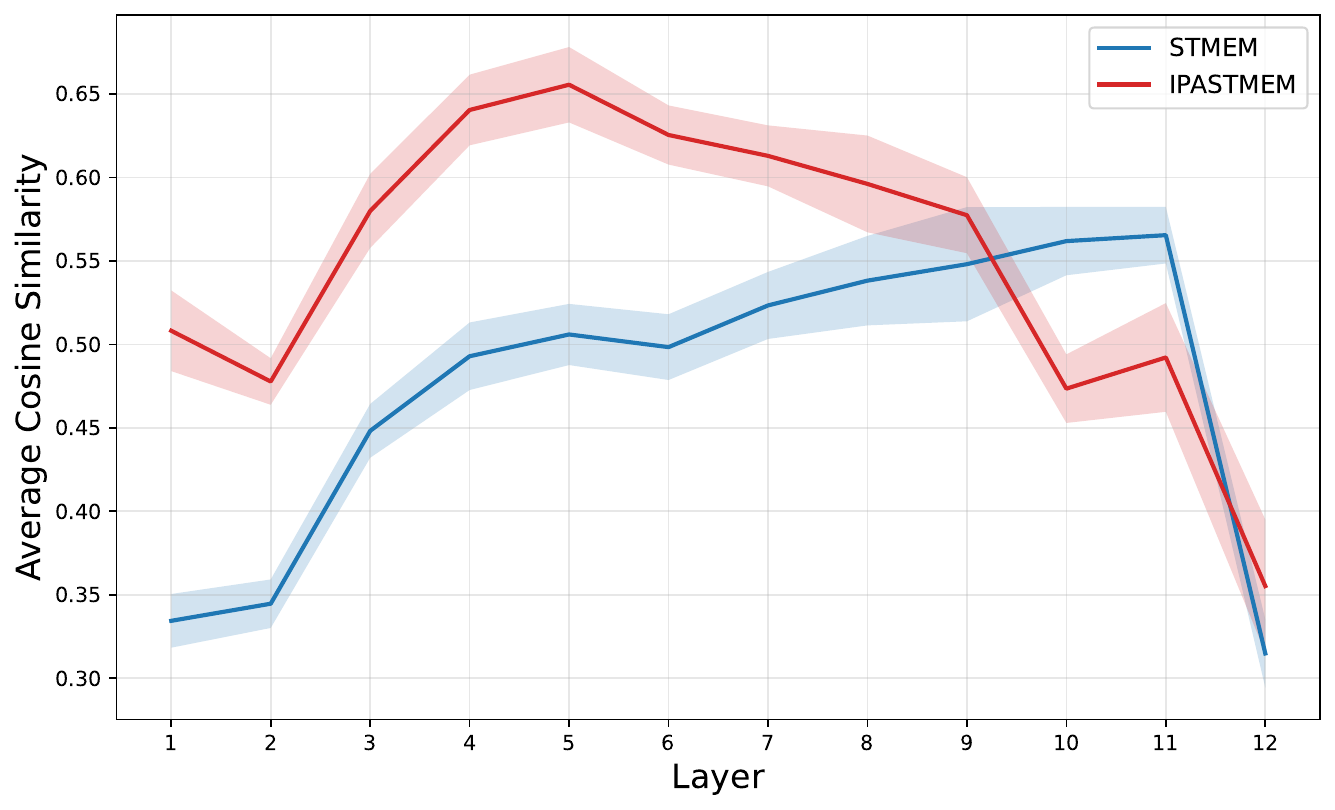}
    \caption{Average cosine similarity through inner layers of pretrained ViT models.}
    \label{fig:average_cosine}
\end{figure}

Figure \ref{fig:average_cosine} illustrates the average cosine similarity between patches across hidden layers of the pretrained models (STMEM and IPASTMEM), computed as the mean across 12 ECG leads, and the standard deviation that captures inter-lead variability. The results reveal a characteristic trajectory: cosine similarity within the same lead gradually increases in the early layers, peaks in the middle layers, and decreases toward the final layers. This trend reflects a progressive synthesis of local information into more homogeneous representations, consistent with prior observations on representation convergence in Transformers \cite{peihao2022, tam2023}. Such a phenomenon is useful in representing the spatial and temporal information of ECG signals. However, as highlighted in \cite{peihao2022, tam2023}, excessive smoothing can occur if not properly controlled, leading to overly uniform embeddings and diminished discriminative power. Fortunately, in this context, the reconstruction loss employed during pretraining is crucial in preventing representation saturation, thereby preserving fine-grained patterns in the last layers and enhancing the ability to capture spatio-temporal dependencies intrinsic to ECG signals. The clear separation of representations in the final layers further enables the model to attend to localized features essential for clinically meaningful ECG interpretation.

\begin{figure}[!t]
  \centering
  \subfloat[STMEM]{%
    \includegraphics[width=0.85\linewidth]{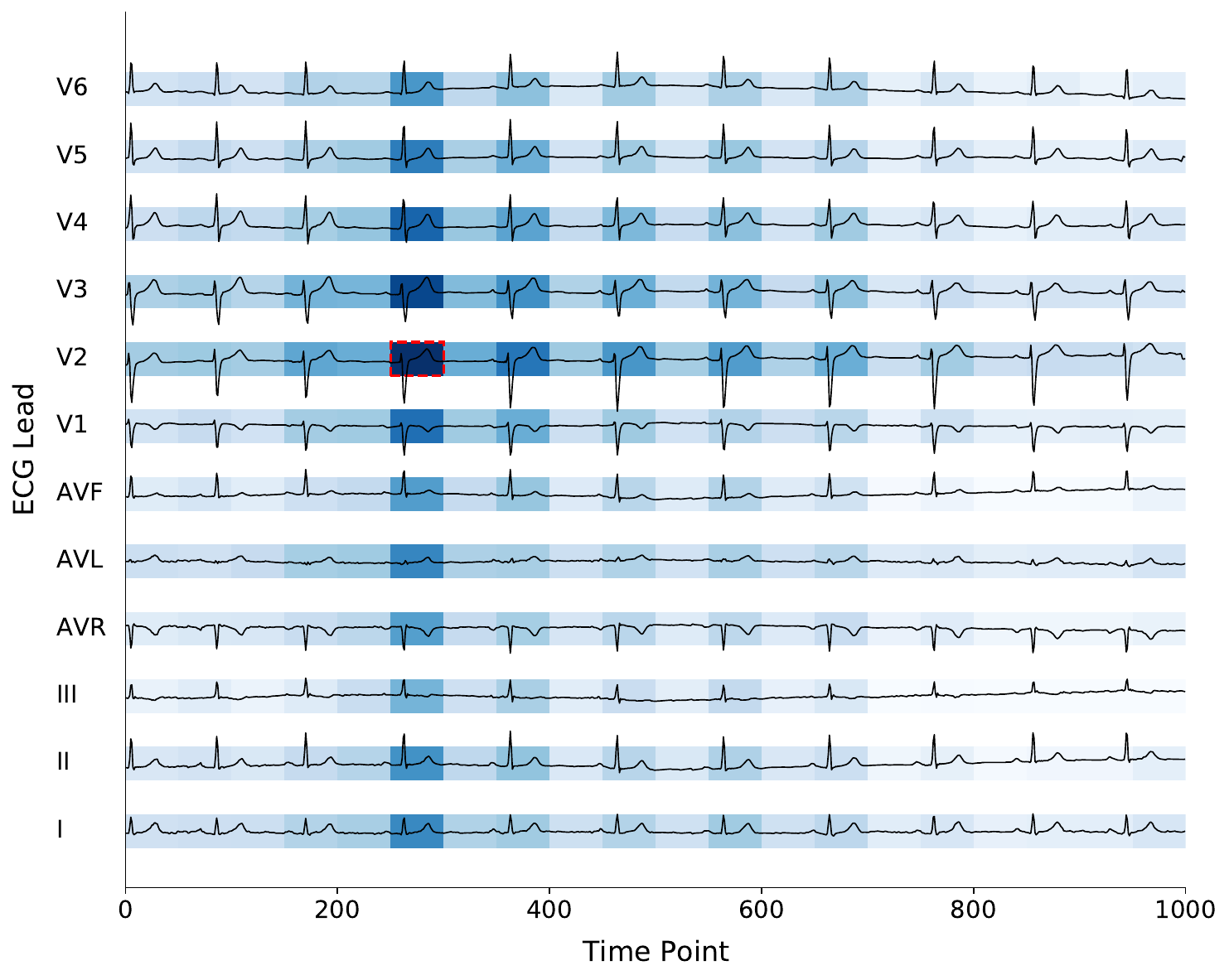}%
    \label{fig:cosmap_stmem}
  }
  \par\medskip
  \subfloat[IPASTMEM]{%
    \includegraphics[width=0.85\linewidth]{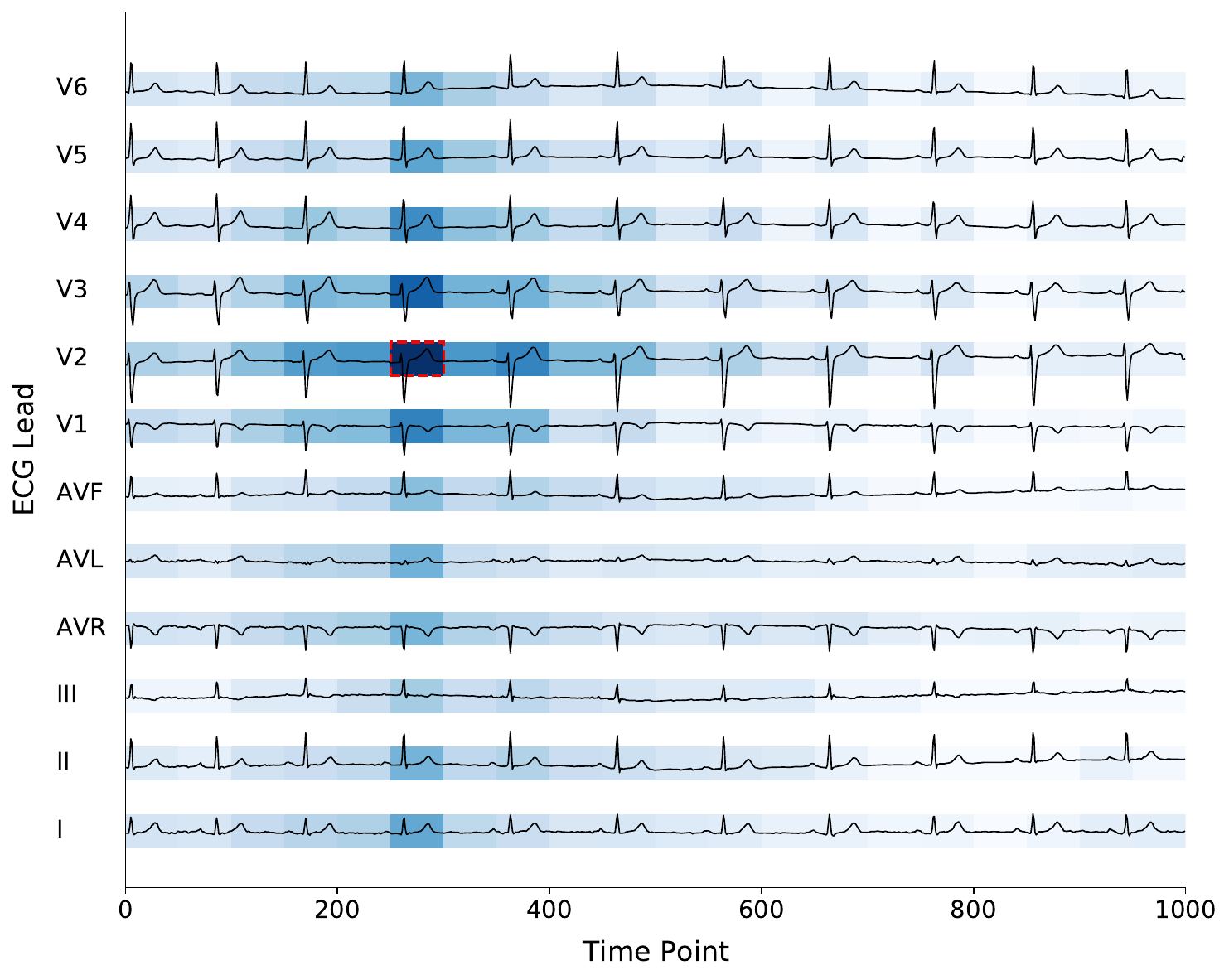}%
    \label{fig:cosmap_ipastmem}
  }

  \caption{Cosine similarity maps of 12-lead ECG provide whole spatial and temporal information regarding the heart: precordial leads (V1-V6) and limb leads (I, II, III, AVR, AVL, AVF). The above figure shows cosine similarity maps for a query patch (i.e., red dashed box) in lead V2 and the remaining patches in two pretrained models: (a) STMEM and (b) IPASTMEM.}
  \label{fig:cosine_maps}
\end{figure}

The comparison between the two plots in Figure \ref{fig:average_cosine} reveals a significant difference in the representation organization. STMEM exhibits a gradual convergence process, with patch similarity steadily increasing across layers before dropping sharply in the final layers. IPASTMEM achieves a high similarity early on (layers 4–5) and then gradually declines. This suggests that IPASTMEM tends to “compress” information at the beginning of the network, then re-establishes representational diversity in later layers. Such behavior arises in IPASTMEM because all layers contribute equally during pretraining, with the early and middle layers also actively involved in the reconstruction process. Notably, these properties are consistently maintained across all ECG leads, confirming that the model's representation learning mechanism does not depend on the individual characteristics of each signal channel but instead exhibits a generalization tendency across the entire input data space. Thus, pretrained ViT models in multi-lead ECG signal processing demonstrate high stability and adaptability. The results of the average cosine similarity across layers explain the phenomenon observed in the linear probing evaluation across layers: \textit{the middle layers often offer superior performance for downstream tasks}. The reason is that the model reaches an optimal information synthesis state at this stage, in which the representations effectively encode the specific spatio-temporal relationship of the ECG signal.

Figure \ref{fig:cosine_maps} visualizes cosine similarity maps between a query patch and other remaining patches in the representation layer of the ViT encoders pretrained on a 12-lead electrocardiogram (ECG) signal. The input signal is 10 seconds long and divided into 20 patches containing 50 data points. In the illustration, a patch on lead V2, marked with a red border, is selected as the query, and the maps represent the cosine correlation between this patch and all other patches on all leads. Regarding space, the highly correlated regions are mainly concentrated in the precordial leads, specifically V1 to V6, which can be explained by the anatomical location of the query patch in lead V2, allowing the model to exploit features within the same anatomical structure preferentially. This result confirms the ability of pretrained models to learn and represent spatial anatomical relationships between leads, even when the input signal is defined as discrete patches. Regarding time, highly correlated patches with the query often exhibit similar ECG waveforms, reflecting the periodicity of physiological signals, a crucial factor in clinical diagnosis. In addition, it is observed that there is a clear difference between the two training strategies, with STMEM exhibiting a higher generalization ability across a wide range of correlations throughout the entire signal, both in space and time. In contrast, IPASTMEM primarily focuses on exploiting information from leads belonging to the same group as the query patch and is less spread out than STMEM, thus enhancing the ability to identify distinct fine-grained features in specific anatomical regions. These results support the hypothesis that ViT can learn complex spatio-temporal relationships in ECG signals and demonstrate that attention representations in the model's hidden layers can provide valuable explanatory information, making an essential contribution to developing explainable deep learning systems in the biomedical domain.

\subsection{Average attention entropy:}

\begin{figure}[H]
    \centering
    \includegraphics[width=8.9cm,height=6.3cm]{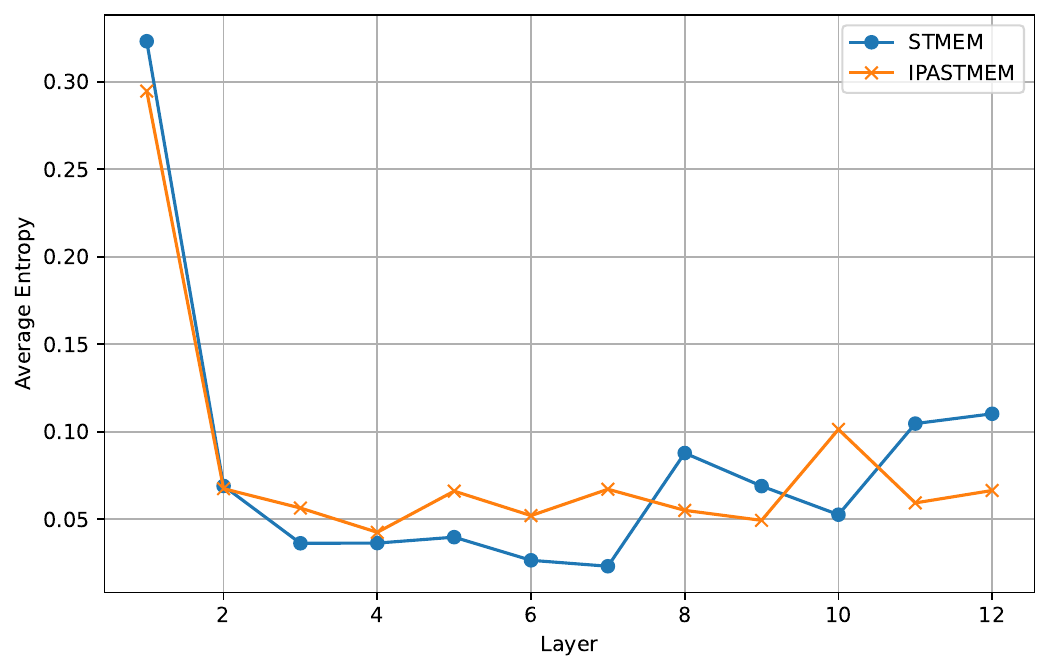}
    \caption{Average attention entropy through inner layers of pretrained ViT models.}
    \label{fig:attention_entropy}
\end{figure}

Analyzing the average attention entropy (AAE) across layers, as shown in Figure \ref{fig:attention_entropy}, provides additional evidence to explain the performance degradation on downstream classification tasks when using the representation from the last layer. Both models exhibit very high entropy in the first layer, reflecting the dispersion of attention and the high uncertainty about the input information. This is typical of the early stages of representation learning when the model has not yet identified the critical regions in the signal. From the second layer onward, the entropy drops rapidly. It stabilizes at a low level, indicating that the model begins to focus on meaningful information and the representation becomes more certain. In the final layers, the entropy increases slightly again, indicating that the model pays more attention to the fine-grained features in the ECG signal. However, the entropy level at the last layer is still significantly lower than at the first. When these observations are compared with the results from linear probing and cosine similarity, it can be concluded that the final layers are not a crude representation but rather a reconstruction stage designed to focus on fine-grained features. Consequently, performance at the final layers tends to decline compared to the middle layers, yet remains substantially higher than at the early stages. The mathematical definition of AAE is provided in Equation \ref{AAE:1}, \ref{AAE:2}.

\noindent Entropy of a single attention vector $\textbf{a}_i^{(h)}$ is defined as
\begin{align}
\label{AAE:1}
H(\textbf{a}_i^{(h)}) = - \sum_{j=1}^N \textbf{a}_{ij}^{(h)} \log \textbf{a}_{ij}^{(h)},
\end{align}
where $H$ is the number of attention heads.

\noindent Average attention entropy (AAE), computed across all tokens and heads, is given by
\begin{align}
\label{AAE:2}
\text{AAE} = \frac{1}{H \cdot N \cdot \log N} \sum_{h=1}^{H} \sum_{i=1}^{N} \sum_{j=1}^N - \textbf{a}_{ij}^{(h)} \log \textbf{a}_{ij}^{(h)}
\end{align}

\section{Experiments and Results}
We evaluated the ECG representations learned by the various methods using two popular strategies: linear probing and fine-tuning, applied in both in-distribution and OOD settings  as in \cite{temesgen2022}. Experimental analyses demonstrated that the proposed methods significantly improve most evaluation metrics, indicating superior performance to the baselines. 

\subsection{In-distribution evaluation:}
The results in Table \ref{tab:ptb} show that, despite only using the linear probing setup, the proposed methods still outperform all the baseline models in both condition and rhythm classification tasks on PTB-XL. The popular SSL methods, such as SimCLR and CPC, perform worse than supervised models. SimCLR degrades sharply in most metrics, indicating that the representations it learns are not rich enough to support downstream tasks effectively. In contrast, the pretrained ViT group shows a clear advantage; here, STMEM achieves comparable performance or even outperforms some supervised models in several metrics such as macro AUC, sample AUC, and sample accuracy. The proposed methods, which incorporate a multi-layer representation fusion mechanism, continue to show superiority. Specifically, PMA (Scheme II) achieves the highest results on the conditional classification problem with approximately 93.44\% macro AUC, 97.22\% sample AUC, 24.73\% macro F1, and 69.93\% sample F1. In comparison, IPASTMEM (Scheme III) stands out in instance accuracy (35.74\%) and sample accuracy (97.89\%). On the rhythm classification problem, PMA continues to lead in sample accuracy (98.32\%) and macro F1 (48.21\%), while IPASTMEM achieves outstanding results in instance accuracy (86.34\%) and sample F1 (86.61\%). These results show that the proposed methods learn significantly more informative and stable representations than the baseline models, achieving superior performance even when updating only the classifier. 

The trends observed in Table \ref{tab:ptb} continue to be consistently maintained in the Chapman dataset, as shown in Table \ref{tab:chapman}, demonstrating the stable generalizability of the proposed methods when moving to another dataset with a smaller scale. In this context, PMA (Scheme II) continues to show a clear advantage, achieving the highest performance in most metrics on the condition classification problem. On the rhythm classification problem, PMA outperforms all the baselines in metrics such as sample AUC (97.88\%), macro F1 (59.33\%), and sample F1 (92.24\%), reflecting the ability to maintain an informative and stable representation even when the training data is of limited size. 

Overall results in in-distribution settings on both PTB-XL and Chapman datasets show that the proposed models consistently outperform the baseline methods, including supervised and self-supervised ones. This reflects the strong representation learning and improved generalization capabilities of the proposed methods, especially in imbalanced multi-label classification, which is common in ECG signals. Among the proposed configurations, Scheme II often outperforms or is par with Scheme I on most metrics and datasets. While Scheme I still provides high and consistent performance, Scheme II tends to exploit deep representations more effectively. This trend is consistent across both the condition and rhythm tasks and across both large (PTB-XL) and smaller (Chapman) data scales, demonstrating the advantage of the PMA compared to PPA.

\renewcommand{\arraystretch}{1.6}
\begin{table*}[htbp]
\centering
\caption{PTB-XL results under linear probing on condition and rhythm classification}
\label{tab:ptb}
\vspace{-0.5em} 
\scriptsize
\resizebox{0.9\textwidth}{!}{
\begin{tabular}{|l|l|l|c|c|c|c|c|c|}
\hline
\textbf{Downstream task} & \textbf{Method} & \textbf{Model} & \multicolumn{6}{c|}{\textbf{Metrics}} \\
\hline
 & & & \textbf{Macro AUC} & \textbf{Sample AUC} & \textbf{Instance Accuracy} & \textbf{Sample Accuracy} & \textbf{Macro F1 Score} & \textbf{Sample F1 Score} \\
\hline
\multirow{9}{*}{Condition}
&\multirow{3}{*}{Supervised}
 & XResnet1d50 \cite{nils2021, temesgen2022} & 90.46$\pm$0.38 & 96.36$\pm$0.13 & 34.96$\pm$1.13 & 97.79$\pm$0.03 & 21.83$\pm$0.86 & 68.94$\pm$0.65 \\
 & & ViT1d \cite{alexey2021, yeongyeon2024} & 85.31$\pm$0.45 & 94.79$\pm$0.24 & 31.27$\pm$1.9 & 97.45$\pm$0.05 & 17.47$\pm$1.6 & 63.79$\pm$1.03 \\
 & & 4FC+2LSTM+2FC \cite{temesgen2022} & 91.14$\pm$0.48 & 96.59$\pm$0.25 & 35.62$\pm$0.93 & 97.8$\pm$0.06 & 23.9$\pm$1.32 & 69.65$\pm$0.8 \\
\cline{2-9}
&\multirow{3}{*}{Self-supervised} 
 & SimCLR \cite{ting2020, temesgen2022} & 84.97$\pm$0.28 & 94.63$\pm$0.31 & 29.37$\pm$1.47 & 97.45$\pm$0.02 & 12.43$\pm$0.62 & 61.88$\pm$0.84 \\
 & & CPC \cite{temesgen2022} & 89.97$\pm$0.3 & 96.04$\pm$0.09 & 32.6$\pm$0.39 & 97.66$\pm$0.01 & 17.12$\pm$0.69 & 66.23$\pm$0.2 \\
 & & STMEM \cite{yeongyeon2024} & 92.6$\pm$0.17 & 96.92$\pm$0.07 & 34.78$\pm$0.6 & 97.82 +/-0.01 & 21.6$\pm$0.9 & 68.41$\pm$0.35 \\
\cline{2-9}
&\multirow{3}{*}{Proposed method}
 & Scheme I - PPA & 93.39$\pm$0.29 & 97.21$\pm$0.09 & 35.42$\pm$0.41 & 97.88$\pm$0.02 & 24.43$\pm$1.41 & 69.82$\pm$0.37 \\
 & & Scheme II - PMA & \textbf{93.44$\pm$0.14} & \textbf{97.22$\pm$0.08} & 35.57$\pm$0.34 & 97.89$\pm$0.02 & \textbf{24.73$\pm$0.78} & \textbf{69.93$\pm$0.31} \\
 & & Scheme III - IPASTMEM & 93.2$\pm$0.3 & 97.14$\pm$0.09 & \textbf{35.74$\pm$0.42} & \textbf{97.89$\pm$0.02} & 23.53$\pm$1.11 & 69.93$\pm$0.33 \\
\hline
\multirow{9}{*}{Rhythm}
&\multirow{3}{*}{Supervised}
 & XResnet1d50 \cite{nils2021, temesgen2022} & 90.13$\pm$3.12 & 94.75$\pm$0.39 & 84.47$\pm$0.46 & 98.1$\pm$0.06 & 40.9$\pm$2.7 & 84.68$\pm$0.54  \\
 & & ViT1d \cite{alexey2021, yeongyeon2024} & 88.37$\pm$0.81 & 93.14$\pm$0.66 & 78.9$\pm$0.96 & 97.28$\pm$0.09 & 32.52$\pm$3.43 & 78.94$\pm$1.05 \\
 & & 4FC+2LSTM+2FC \cite{temesgen2022} & 93.94$\pm$3 & 91.72$\pm$7.84 & 78.6$\pm$14.07 & 97.59$\pm$1.12 & 37.4$\pm$8.93 & 78.43$\pm$14.87 \\
\cline{2-9}
&\multirow{3}{*}{Self-supervised} 
 & SimCLR \cite{ting2020, temesgen2022} & 87.12$\pm$0.28 & \textbf{96.28$\pm$0.48} & 81.8$\pm$0.36 & 97.43$\pm$0.04 & 25.82$\pm$1.39 & 83.77$\pm$0.63 \\
 & & CPC \cite{temesgen2022} & 91.51$\pm$0.37 & 94.33$\pm$0.17 & 82.05$\pm$0.23 & 97.74$\pm$0.03 & 30.69$\pm$2.58 & 82.25$\pm$0.27 \\
 & & STMEM \cite{yeongyeon2024} & \textbf{97.25$\pm$0.21} & 95.33$\pm$0.12 & 85.55$\pm$0.38 & 98.24$\pm$0.05 & 44.1$\pm$2.73 & 85.72$\pm$0.28 \\
\cline{2-9}
&\multirow{3}{*}{Proposed method}
 & Scheme I - PPA & 97.18$\pm$0.16 & 95.5$\pm$0.12 & 86.02$\pm$0.24 & 98.3$\pm$0.03 & 46.79$\pm$2.73 & 86.28$\pm$0.24 \\
 & & Scheme II - PMA & 97.14$\pm$0.24 & 95.68$\pm$0.11 & 86.27$\pm$0.26 & \textbf{98.32$\pm$0.04} & \textbf{48.21$\pm$4.17} & 86.6$\pm$0.15 \\
 & & Scheme III - IPASTMEM & 96.66$\pm$0.16 & 95.7$\pm$0.12 & \textbf{86.34$\pm$0.28} & 98.31$\pm$0.03 & 47.46$\pm$2.24 & \textbf{86.61$\pm$0.25} \\
\hline
\end{tabular}
}
\end{table*}
\renewcommand{\arraystretch}{1.6}
\begin{table*}[htbp]
\centering
\caption{Chapman results under linear probing on condition and rhythm classification}
\label{tab:chapman}
\vspace{-0.5em} 
\scriptsize
\resizebox{0.9\textwidth}{!}{
\begin{tabular}{|l|l|l|c|c|c|c|c|c|}
\hline
\textbf{Downstream task} & \textbf{Method} & \textbf{Model} & \multicolumn{6}{c|}{\textbf{Metrics}} \\
\hline
 & & & \textbf{Macro AUC} & \textbf{Sample AUC} & \textbf{Instance Accuracy} & \textbf{Sample Accuracy} & \textbf{Macro F1 Score} & \textbf{Sample F1 Score} \\
\hline
\multirow{9}{*}{Condition}
&\multirow{3}{*}{Supervised}
 & XResnet1d50 \cite{nils2021, temesgen2022} & 82.97$\pm$0.87 & 96.68$\pm$0.58 & 49.54$\pm$1.55 & 98.62$\pm$0.04 & 19.44$\pm$1.56 & 75.25$\pm$1.17 \\
 & & ViT1d \cite{alexey2021, yeongyeon2024} & 76.97$\pm$0.7 & 93.78$\pm$1.35 & 42.13$\pm$1.72 & 98.22$\pm$0.06 & 12.85$\pm$1.04 & 64.69$\pm$2.3 \\
 & & 4FC+2LSTM+2FC \cite{temesgen2022} & 82.67$\pm$2.17 & 95.63$\pm$3.86 & 45.71$\pm$14.66 & 98.28$\pm$1.07 & 19.2$\pm$4.64 & 70.5$\pm$14.86 \\
\cline{2-9}
&\multirow{3}{*}{Self-supervised} 
 & SimCLR \cite{ting2020, temesgen2022} & 77.37$\pm$0.35 & 89.44$\pm$0.62 & 36.35$\pm$0.72 & 98.17$\pm$0.02 & 12.52$\pm$0.95 & 57.81$\pm$0.95 \\
 & & CPC \cite{temesgen2022} & 81.96$\pm$0.48 & 95.19$\pm$0.26 & 47.28$\pm$0.63 & 98.51$\pm$0.02 & 14.87$\pm$0.68 & 70.4$\pm$0.47 \\
 & & STMEM \cite{yeongyeon2024} & 85.24$\pm$0.22 & 97.76$\pm$0.24 & 53.58$\pm$0.66 & 98.77$\pm$0.01 & 19.76$\pm$1.12 & 78.5$\pm$0.52 \\
\cline{2-9}
&\multirow{3}{*}{Proposed method}
 & Scheme I - PPA & 85.63$\pm$0.28 & 98.16$\pm$0.19 & 55.41$\pm$0.54 & 98.85$\pm$0.01 & \textbf{21.45$\pm$0.45} & 80.19$\pm$0.3 \\
 & & Scheme II - PMA & 85.69$\pm$0.3 & \textbf{98.18$\pm$0.23} & \textbf{55.43$\pm$0.38} & \textbf{98.85$\pm$0.01} & 21.29$\pm$0.52 & \textbf{80.25$\pm$0.38} \\
 & & Scheme III - IPASTMEM & \textbf{85.72$\pm$0.3} & 97.91$\pm$0.25 & 54.4$\pm$0.66 & 98.82$\pm$0.02 & 20.72$\pm$1.3 & 79.29$\pm$0.61 \\
\hline
\multirow{9}{*}{Rhythm}
&\multirow{3}{*}{Supervised}
 & XResnet1d50 \cite{nils2021, temesgen2022} & 91.94$\pm$1.5 & 96.06$\pm$0.62 & 86.95$\pm$1.71 & 98.22$\pm$0.18 & 52.8$\pm$1.63 & 87.75$\pm$1.41  \\
 & & ViT1d \cite{alexey2021, yeongyeon2024} & 88.26$\pm$2.78 & 92.77$\pm$0.61 & 75.6$\pm$1.48 & 96.44$\pm$0.2 & 40.12$\pm$2.11 & 76.36$\pm$1.39 \\
 & & 4FC+2LSTM+2FC \cite{temesgen2022} & 92.96$\pm$1.25 & 97.08$\pm$0.53 & 88.17$\pm$1.85 & 98.26$\pm$0.24 & 56.37$\pm$1.7 & 88.81$\pm$1.66 \\
\cline{2-9}
&\multirow{3}{*}{Self-supervised} 
 & SimCLR \cite{ting2020, temesgen2022} & 83.31$\pm$0.24 & 86.64$\pm$0.59 & 61.96$\pm$1.24 & 95.55$\pm$0.09 & 37.28$\pm$1.52 & 64.88$\pm$1.07 \\
 & & CPC \cite{temesgen2022} & 91.47$\pm$0.36 & 94.03$\pm$0.62 & 80.18$\pm$1.25 & 97.32$\pm$0.11 & 45.81$\pm$0.7 & 81.2$\pm$1.12 \\
 & & STMEM \cite{yeongyeon2024} & 94.42$\pm$0.07 & 97.6$\pm$0.16 & 90.44$\pm$0.47 & 98.67$\pm$0.06 & 57.99$\pm$0.88 & 91.16$\pm$0.29 \\
\cline{2-9}
&\multirow{3}{*}{Proposed method}
 & Scheme I - PPA & \textbf{94.68$\pm$0.1} & 97.85$\pm$0.22 & \textbf{91.79$\pm$0.58} & \textbf{98.85$\pm$0.06} & 59.19$\pm$1.19 & 92.23$\pm$0.39 \\
 & & Scheme II - PMA & 94.65$\pm$0.07 & \textbf{97.88$\pm$0.22} & 91.71$\pm$0.55 & 98.84$\pm$0.06 & \textbf{59.33$\pm$1.12} & \textbf{92.24$\pm$0.42} \\
 & & Scheme III - IPASTMEM & 94.67$\pm$0.11 & 97.82$\pm$0.23 & 91.51$\pm$0.39 & 98.8$\pm$0.04 & 58.77$\pm$0.76 & 91.89$\pm$0.45 \\
\hline
\end{tabular}
}
\end{table*}

\subsection{Out-of-distribution evaluation:}
When switching to the OOD evaluation setting in Table \ref{tab:ptb_out_dis} (Appendix \ref{Appendix:D}), where the entire PTB-XL dataset is not used in the pretraining phase, the observed trends in the in-distribution remain consistent. In this setting, IPASTMEM (Scheme III) continues to demonstrate consistent and superior performance, achieving the highest values on most metrics in the PTB-XL condition classification task, while PPA and PMA also maintain superior results compared to the remaining baselines. For the rhythm classification task, IPASTMEM achieves 97.02\% macro AUC and 95.56\% sample AUC — two key metrics directly reflecting the ability to discriminate rare disease classes, which are often significantly impaired in imbalanced datasets. Maintaining nearly constant performance when moving from in-distribution to OOD shows that the learned vector representations are rich in information and have strong generalization ability, helping the model perform stably even when the data distribution changes radically.

\subsection{Ablation Study}
The combined analysis from Tables \ref{tab:ptb_gap}, \ref{tab:chapman-gap}, and \ref{tab:ptbxl-gap-ood} (Appendix \ref{Appendix:E}) shows a clear difference in the performance gap between linear probing and full model fine-tuning across the method groups. SimCLR consistently exhibits a huge gap: relatively low performance with linear probing but a sharp increase with fine-tuning on both PTB-XL (in-distribution and OOD) and Chapman, reflecting that the representations learned from SimCLR are not informative enough and rely almost entirely on full parameter updates to achieve high performance. In contrast, CPC and STMEM exhibit only a small gap, indicating stable representation quality and better generalization ability. In particular, the proposed methods maintain a small gap between the two settings and, in some cases, slightly degrade performance, indicating that they have learned informative and stable representations robust enough to achieve high performance with linear probing alone. However, when applying fine-tuning, the proposed methods maintain superior performance over the entire baseline, while the observed characteristics in Tables \ref{tab:ptb}, \ref{tab:chapman}, and Table \ref{tab:ptb_out_dis} in Appendix \ref{Appendix:D} (linear probing) remain unchanged.

Another observation is that although our proposed methods generally outperform SimCLR in various datasets and settings, the PTB-XL rhythm classification results show that SimCLR outperforms the proposed method in some metrics, if fine-tuning. This phenomenon can be explained by the characteristics of the problem and the data. First, rhythm classification is a relatively simple task with only 11 classes, while the PTB-XL dataset is large enough to fine-tune the ResNet backbone effectively from pretrained weights. In such a context, fine-tuning helps the ResNet in SimCLR adjust the weights and exploit the full power of the network architecture, thereby improving the performance, even surpassing the proposed methods in some cases. However, the linear probing section observation shows that SimCLR is inferior to the proposed methods, reflecting that the representation learned from SimCLR is not rich in information for the downstream task. SimCLR improves only with fine-tuning, mainly due to the ability to update all ResNet weights on a simple task with large enough data, rather than the inherent quality of pretraining. In contrast, in more complex settings, such as PTB-XL condition classification, or on smaller datasets, such as Chapman, SimCLR significantly underperforms compared to the proposed methods. Thus, this result confirms the limitations of SimCLR in learning general representations and reinforces the superiority of our proposed methods in maintaining stable performance across different contexts, especially complex tasks or limited data, which are more common in biomedical practice.

In summary, essential conclusions can be drawn from the above results. 
\begin{itemize}
\item Across both evaluation scenarios, in-distribution and OOD, self-supervised learning methods significantly improve over supervised learning methods on most evaluation metrics, classification tasks, and datasets.
\item Our proposed methods consistently outperform baselines - including self-supervised and supervised models - with stable performance across various datasets, classification tasks, and evaluation criteria.
\item When applying pretrained ViT to downstream classification tasks, combining information from multiple layers improves the representation quality, in which the PMA method outperforms PPA.
\item Comparing the two pretrained strategies, STMEM and IPASTMEM, shows that integrating layers in the encoder of ViT significantly improves the model's efficiency on OOD datasets.
\end{itemize}

\section{Conclusion}
In this study, we comprehensively analyzed the impact of each layer within the pretrained ViT on ECG signals. We demonstrated that relying solely on the last layer, which is common in practice, does not provide optimal performance. Through experiments on various datasets, evaluation metrics, and downstream tasks, we observed a consistent pattern: the performance of the early layers is typically lowest, gradually increases and peaks in the middle layers, and then slightly decreases in the last layers. Based on this finding, we proposed three strategies for exploiting multilayer representations, including (i) Post-pretraining Pooling-based Aggregation (PPA), (ii) Post-pretraining Mixture-of-layers Aggregation (PMA), and (iii) In-pretraining Pooling-based Aggregation STMEM (IPASTMEM) to enhance the quality of the base representation. Experimental results have demonstrated that all three methods improve generalization and deliver superior performance, especially in non-distributional data, thereby highlighting the potential of exploiting multi-layer information in pretrained Transformer models for biomedical applications. We plan to extend our research to multimodal ECG–text problems to integrate medical knowledge into the ECG signal representation. This approach aims to improve the model’s ability to understand physiological and pathological context, thereby enabling accurate recognition of labels not included in the training using zero-shot learning on downstream tasks.

\section{Acknowledgements}
This research is partially funded by BF-PhD: "VHeart FM: a foundation model for ECG analysis in Vietnam", partially funded from HORIZON-HLTH-2022-IND-13: "Privacy compliant health data as a service for AI development (PHASE IV AI)", funded by the European Union, under Grant Agreement \#101095384 and from the Flemish Government (AI Research Program). Maarten De Vos and Christos Chatzichristos are affiliated with Leuven.AI - KU Leuven Institute for AI, B-3000, Leuven, Belgium, and are partially funded.

\bibliographystyle{IEEEtran}
\bibliography{Foundation_ECG}
\vspace{-5.9em}
\begin{IEEEbiography}[{\includegraphics[width=1in,height=1.25in,clip,keepaspectratio]{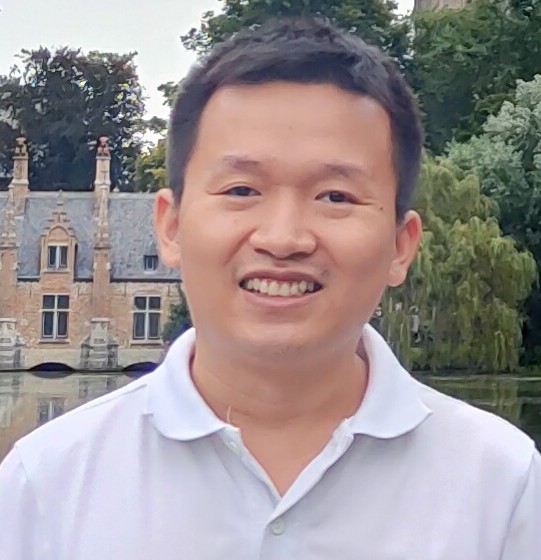}}]
	{Phu X. Nguyen} received the B.E. degree in electronics and telecommunication engineering from Ho Chi Minh City University of Technology, Vietnam, in 2017, and the M.E. degree in electronic engineering from Soongsil University, South Korea, in 2019. He was a machine learning engineer at Cybercore Co., Ltd. (2019–2020) and has been a lecturer at FPT University, Vietnam, since 2020. From 2021 to 2023, he was a senior research engineer at NextG, FPT AI. He is currently pursuing the Ph.D. degree at the Department of Electrical Engineering, KU Leuven, Belgium. His research focuses on machine learning, statistical learning, and optimization, with applications in IoT, audio/speech, and biosignal analysis. In 2025, he was part of the first-place winning team in the George B. Moody PhysioNet Challenge for Chagas disease detection from ECG signals. 
\end{IEEEbiography}
\biogap
\begin{IEEEbiography}[{\includegraphics[width=1in,height=1.25in,clip,keepaspectratio]{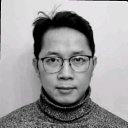}}]
	{Huy Phan} received the M.Eng. degree from Nanyang Technological University, Singapore, in 2012, and the Dr.-Ing. degree in computer science from University of L\"ubeck, Germany, in 2017. From 2017 to 2018, he was a Postdoctoral Research Assistant with University of Oxford, UK. From 2019 to 2020, he was a Lecturer at University of Kent, UK. From 2020 to 2022, he was a Lecturer in AI at Queen Mary University of London and Turing Fellow at the Alan Turing Institute, London, UK. From 2023-2024, he was a senior research scientist at Amazon AGI, Cambridge, USA. In November 2024, he joined Meta Reality Labs in Paris, France where he is a research scientist. His research interests include machine learning and signal processing with a focus on audio/speech and biosignal analysis. In 2018, he received the Bernd Fischer Award for the best PhD thesis from University of L\"ubeck. In 2021, he was awarded Benelux’s IEEE-EMBS Best Paper Award 2019-20.  
\end{IEEEbiography}
\biogap
\begin{IEEEbiography}[{\includegraphics[width=1in,height=1.25in,clip,keepaspectratio]{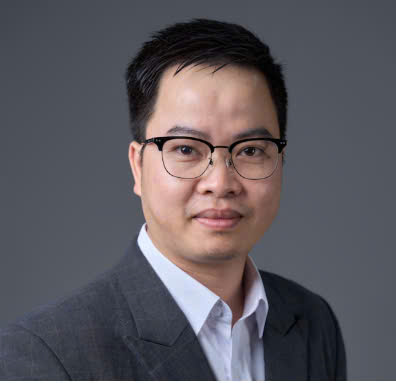}}]
	{Hieu Pham} is an Assistant Professor at the College of Engineering and Computer Science (CECS), VinUniversity, and a Principal Investigator at VinUni-Illinois Smart Health Center. He received his Ph.D. in Computer Science from the Toulouse Computer Science Research Institute (IRIT), University of Toulouse, France, in 2019 and joined the Coordinated Science Laboratory at the University of Illinois Urbana-Champaign (UIUC), USA, as a Visiting Scholar in 2023. Previously, he earned a Degree of Engineer in Industrial Informatics from Hanoi University of Science and Technology (HUST), Vietnam, in 2016. His research includes Artificial Intelligence (AI), Machine Learning, Deep Learning, and Computer Vision, especially their applications in Smart Healthcare, e.g., Medical Imaging Diagnosis, AI-based Computer-aided Diagnosis (AI-CAD), AI-assisted Diagnosis and Treatment, AI-assisted Disease Prevention and Risk Monitoring.
\end{IEEEbiography}
\biogap
\begin{IEEEbiography}[{\includegraphics[width=1in,height=1.25in,clip,keepaspectratio]{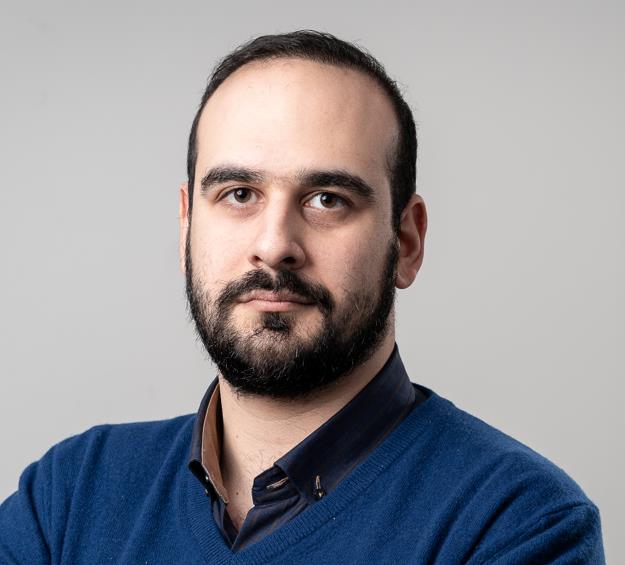}}]
	{Christos Chatzichristos} is a Postdoctoral Researcher affiliated with the department of Electrical Engineering (ESAT) of KU Leuven (KUL) and with VAIA (Flemish AI academy), as an AI-expert in the field of healthcare.  He gained his PhD in 2019 from the Department of Informatics and Telecommunications, Un. of Athens and was awarded a Marie Curie Skolodowska fellowship for the completion of his PhD research. Christos obtained a MSc in Biomedical Engineering from KUL, and a Diploma in Electrical and Computer engineering from AUTH. He has been an author of multiple peer-reviewed papers He has been the first author of a paper received the best paper award in IEEE SPMB 2020, and a member of the team that won the first price in the Neureka Challenge 2020 for seizure detection. 
\end{IEEEbiography}
\biogap
\begin{IEEEbiography}[{\includegraphics[width=1in,height=1.25in,clip,keepaspectratio]{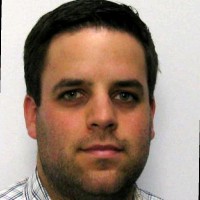}}]
	{Bert Vandenberk} received the M.D. and Ph.D. degrees in Cardiology from KU Leuven, Belgium, and the M.Sc. degree in Clinical Trials from the University of London, U.K. He is currently an Assistant Professor with the Department of Cardiovascular Sciences, KU Leuven, and a cardiac electrophysiologist at UZ Leuven. His research interests include artificial intelligence in cardiology, computational electrocardiology, and data-driven approaches for longitudinal risk stratification and treatment of complex arrhythmias.
\end{IEEEbiography}
\biogap
\begin{IEEEbiography}[{\includegraphics[width=1in,height=1.25in,clip,keepaspectratio]{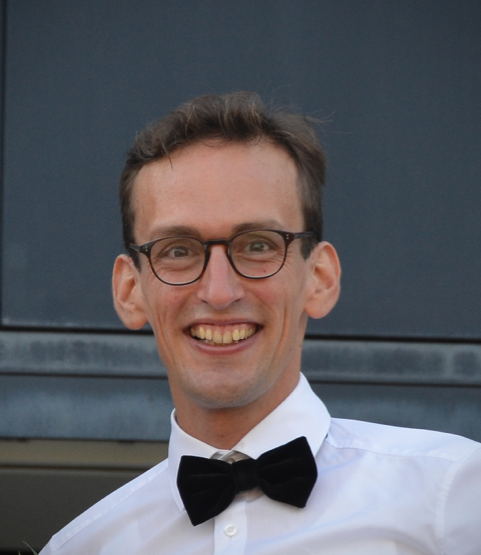}}]
	{Maarten De Vos} is Professor in the Departments of Engineering and Medicine at KU Leuven after being Associate Professor at the University of Oxford (UK) and Junior Professor at the University of Oldenburg (Germany). Since the start of his career, he has focused on improving data science approaches for various healthcare applications.  Currently, his Artificial Intelligence (AI) solutions are used in various hospital departments, ranging from neonatology to elderly care.\\His pioneering research has won innovation prices, among which the prestigious Mobile Brain Body monitoring prize (2017), the Martin Black Prize for the best paper in Physiological Measurements (2019) and in the IEEE EMBS Benelux award for best paper in the biomedical field (2021). He also received the early career prize for his technical contributions in 2023 from KVAB, He has a strong interest in translational research, and advises various healthcare spin-off companies. He is associate editor for IEEE Journal of Biomedical Health Informatics and on the editorial board of Journal of Neural Engineering and he coordinates the online EdX course on AI in healthcare.
\end{IEEEbiography}
\onecolumn
\title{ECG-Soup: Harnessing Multi-Layer Synergy for ECG Foundation Models}
\appendices
\noindent \textbf{SUPPLEMENTARY MATERIAL:}
In this supplementary material, we present the theoretical basis for the operation of the masked vision transformer (ViT) in detail and provide a mathematical proof of Lemma 1. In addition, we extend the analysis with additional results, including hidden layer analysis, out-of-distribution performance, and an ablation study to clarify the reasons for the performance discrepancy between linear probing and fine-tuning.

\section{1D Vision Transformer Backbone for 12-lead ECG}
\label{Appendix:A}
\subsection{1D Vision Transformer Backbone for 12-lead ECG}
\noindent \textbf{Patch Embedding:}
Denote an input ECG signal as ${\bf X} \in \mathbb{R}^{C \times L}$, where $C$ is the number of leads and $L$ is the length of the ECG signal. We split the ECG signal into 1D non-overlapping segments called ECG patches and denoted by

\begin{align}
\label{eq:10}
{\bf X}_{p} \in \mathbb{R}^{C \times N \times P},
\end{align}

\noindent where $N = \left\lfloor \frac{L}{P} \right\rfloor$ is the number of patches and $P$ is the length of a patch, respectively. ECG patches are then projected into $D$-dimensional patch embeddings using a linear projection
:

\begin{align}
\label{eq:11}
{\bf Y}_{0} = {\bf X}_{p}{\bf W}_{e} + {\bf b}_{e},
\end{align}

\noindent where ${\bf Y}_{0} \in \mathbb{R}^{C \times N \times D}$, ${\bf W}_{e} \in \mathbb{R}^{P \times D}$, and ${\bf b}_{e} \in \mathbb{R}^{D}$.

\noindent\textbf{Positional Encoding and Lead Encoding:}
A shared embedding, denoted as $[SEP]$, is inserted before and after the sequence of patch embeddings to support the model in distinguishing between patch embeddings from different leads, ${\bf Y}_{0}$,  resulting in

\begin{align}
\label{eq:12}
\mathbf{Y}^{'}_{0} = [[SEP] \; \; \mathbf{Y}_{0} \; \; [SEP]] \in \mathbb{R}^{C \times (N+2) \times D}. 
\end{align}

\noindent To effectively model the ECG features, we add the learnable positional embeddings to the patch embeddings:

\begin{align}
\label{eq:13}
{\bf Y}_{0}^{''} = {\bf Y}^{'}_{0} + {\bf E}_{pos}. 
\end{align}

\noindent To improve the discriminative capability among leads, the learnable lead embeddings are added to ${\bf Y}_{0}^{''}$ as follows

\begin{align}
\label{eq:14}
{\bf Y} = {\bf Y}^{''}_{0} + {\bf E}_{lead}, 
\end{align}

\noindent \textbf{Transformer Encoder:}
We stack 12 Transformer layers in the encoder, where each layer comprises three components:

\noindent \textit{i) Multi-head Self-Attention (MSA):}

\begin{align}
\label{eq:15}
\mathrm{MSA}(\mathbf{Y}) = \mathrm{Concat}(\mathrm{head}_1, \ldots, \mathrm{head}_h) \mathbf{W}^{O},
\end{align}

\noindent where each head is computed by a softmax function as follows

\begin{align}
\label{eq:16}
\mathrm{head}_i = \mathrm{softmax}\left( \frac{\mathbf{Q}_i \mathbf{K}_i^\top}{\sqrt{d_k}} \right) \mathbf{V}_i,
\end{align}

\noindent where
${\bf Q}_i = {\bf Y}{\bf W}_i^Q$,
${\bf K}_i = {\bf Y}{\bf W}_i^K$,
${\bf V}_i = {\bf Y}{\bf W}_i^V$,
${\bf W}_i^Q, {\bf W}_i^K, {\bf W}_i^V \in \mathbb{R}^{D \times d_k}$,
${\bf W}^{O} \in \mathbb{R}^{h \cdot d_k \times D}$.

\noindent \textit{ii) Add \& Norm:}

\begin{align}
\label{eq:17}
{\bf Y}^{(l)'} = \mathrm{LayerNorm}\left({\bf Y}^{(l-1)} + \mathrm{MSA}\left({\bf Y}^{(l-1)}\right)\right). 
\end{align}

\noindent \textit{iii) Feed Forward Network (FFN):}

\begin{align}
&\mathrm{FFN}(x) = \mathrm{GELU}({\bf x}{\bf W}_1 + {\bf b_1}){\bf W}_2 + {\bf b_2}, \label{eq:18} \\
&{\bf Y}^{(l)} = \mathrm{LayerNorm}\left({\bf Y}^{(l)'} + \mathrm{FFN}({\bf Y}^{(l)'})\right). \label{eq:19}
\end{align}

\subsection{Masked Vision Transformer}
\begin{figure*}[t]
    \centering
    \includegraphics[width=\textwidth]{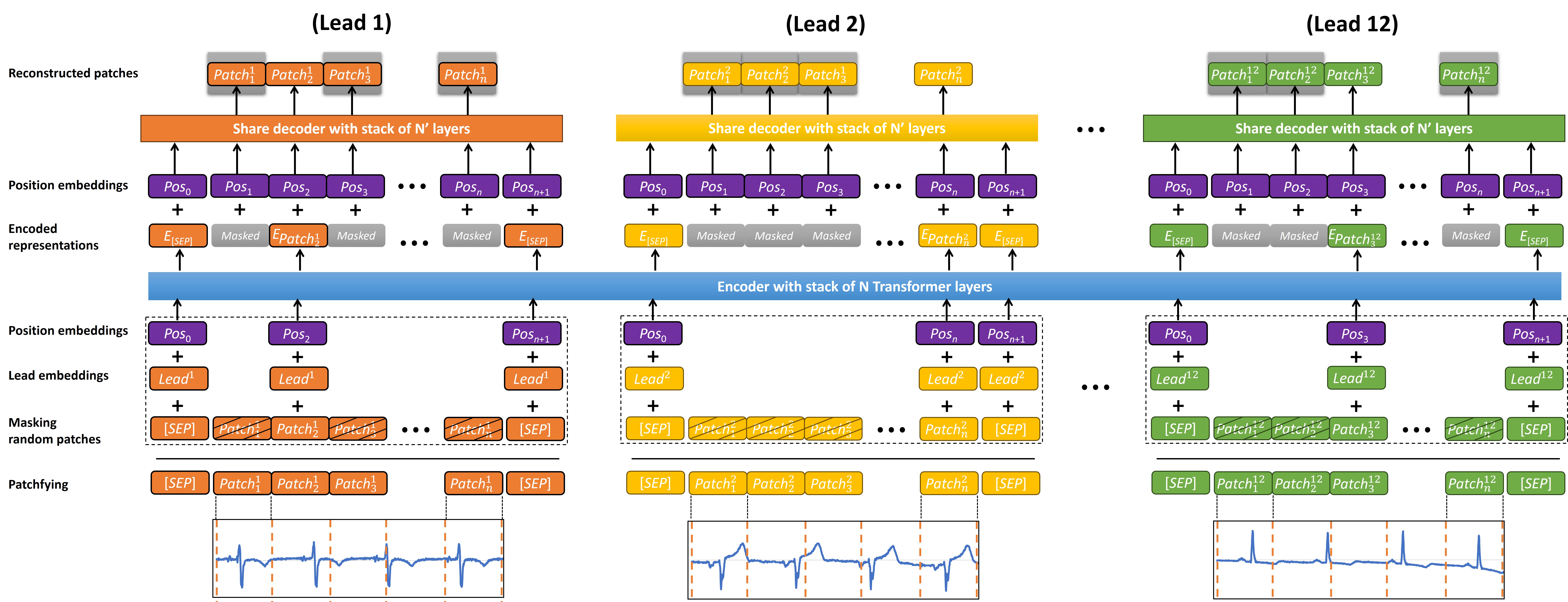}
    \caption{The overview of Spatio-Temporal Masked Electrocardiogram Modeling (STMEM) \cite{yeongyeon2024}.}
    \label{fig:1}
\end{figure*}

The ECG patches are randomly masked in the pre-text task. The unmasked patches are fed into the ViT encoder, while the masked patches are reconstructed through the decoder-based Transformers as illustrated in Figure \ref{fig:1}.

\noindent \textbf{Encoder:}
In the pretraining phase, a random masking strategy is applied on the embedding sequence to reduce the inherent redundant information in the ECG signal and simultaneously avoid overfitting. We denote the unmasked embeddings as $${\bf Y}^{''}_{0-unmask} = \{{\bf Y}^{''}_{0^{(0)}}, {\bf Y}^{''}_{0^{(i_{1})}}, \ldots, {\bf Y}^{''}_{0^{(i_{S})}}, {\bf Y}^{''}_{0^{(N + 1)}}\},$$ and the masked embeddings as $${{\bf Y}^{''}_{0-mask} = \{{\bf Y}^{''}_{0^{(j_{1})}}, \ldots, {\bf Y}^{''}_{0^{(j_{S'})}}\}},$$ where ${\bf Y}^{''}_{0^{(0)}}$ and ${\bf Y}^{''}_{0^{(N + 1)}}$ are embeddings resulted from $[SEP]$ tokens,  $S$ and $S'$ are the number of unmasked and masked embeddings, respectively, with $S + S' = N$ and a masking ratio $m = \dfrac{S'}{N} \in [0, 1]$. Note that $i_{s}$ and $j_{s'}$ are randomly selected from the embedding sequence, not including $[SEP]$ tokens. The unmasked embeddings are input to the Transformer layers for representation learning, whereas the masked embeddings will be employed as the reconstruction targets.

\noindent The learnable lead embeddings are added to the unmasked ECG embeddings as in \eqref{eq:14}:

\begin{align}
\label{eq:20}
{\bf Y}_{unmask} = {\bf Y}^{''}_{0-unmask} + {\bf E}_{lead}. 
\end{align}

\noindent The ECG embeddings are then fed into the stack of Transformer layers to output the encoded embeddings as in \eqref{eq:15}-\eqref{eq:19}

\begin{align}
\label{eq:21}
\overline{{\bf Y}} = \mathrm{Encoder}({\bf Y}_{unmask}). 
\end{align}

\noindent \textbf{Decoder:}
The decoder receives the encoded embeddings. These embeddings are first projected into $D^{'}$-dimension embeddings using linear projection:

\begin{align}
\label{eq:22}
{\bf Z}_{0} = \overline{{\bf Y}}{\bf W}_{d} + {\bf b}_{d},
\end{align}

\noindent where ${\bf W}_{d} \in \mathbb{R}^{D \times D^{'}}$, $\mathbf{Z}_{0} \in \mathbb{R}^{C \times (S + 2) \times D^{'}}$. A learnable shared mask embedding, $\mathbf{E}_{mask} \in  \mathbb{R}^{C \times S^{'} \times D^{'}}$, is then shuffled into ${{\bf Z}_{0}}$ to create an expanding sequence, ${{\bf Z}_{cat}} \in \mathbb{R}^{C \times (N + 2) \times D^{'}}$, and reconstruct the original order of the elements, as in the original ECG embedding. Similarly to the encoder, the learnable positional embeddings are also added to ${\bf Z}_{cat}$ to provide positional information.

\begin{align}
\label{eq:23}
\mathbf{Z} = \mathbf{Z}_{cat} + \mathbf{E}_{decoder-pos}. 
\end{align}

\noindent The embedding sequence, $\mathbf{Z}$, is sent to the shared lead-wise decoder with four additional Transformer blocks to reconstruct the masked segments. The training objective aims to minimize the error between the original ECG signals of the masked segments,  $\{{\bf X}_{i}\}_{i \in \mathcal{M}}$, and their corresponding reconstruction outputed by the decoder,  $\{\hat{{\bf X}}_{i}\}_{i \in \mathcal{M}}$

\begin{align}
\label{eq:24}
\mathcal{L}_{\text{reconst}} = \frac{1}{|\mathcal{M}^{'}|} \sum_{i \in \mathcal{M}^{'}} \left\| \hat{\mathbf{X}}_i - \mathbf{X}_i \right\|_2^2, 
\end{align}

\noindent where $\mathcal{M}^{'}$ is the set of masked locations.

\noindent The shared lead-wise decoder is deliberately designed for multi-lead ECG. The spatio-temporal patchifying enables the decoder to access unmasked embeddings from multiple leads, all aligned with the same temporal information. Thus, the training process may become too simple, resulting in a negative impact on the representation of the encoder. To address this limitation, the decoder is designed to process the embedding sequence from each lead independently. This guarantees that the decoder does not directly exploit information from other leads during the reconstruction process. Such a design choice increases the task's difficulty, encouraging the encoder to learn the spatio-temporal representation more efficiently.

\section{The proof of Lemma 1}
\label{Appendix:B}
For any vector $\textbf{b}_{k} \subset \mathbb{R}^d$ and row $i, j$; the following inequality holds:

\begin{equation} \label{eq:3}
\begin{aligned}
\left\| \sum_{k} a_{ik} \textbf{b}_k - \sum_{k} a_{jk} \textbf{b}_k \right\|
\leq \delta(\textbf{A}) \max_{x,y} \| \textbf{b}_x - \textbf{b}_y \|.
\end{aligned}
\end{equation}

\noindent Let $d_k = \min \{a_{ik}, a_{jk} \}$ and $e = \sum_{k} d_k \quad (0 \leq e \leq 1)$

\noindent The residual of each row

\begin{equation} \label{eq:25}
\begin{aligned}
u_k = a_{ik} - d_k \geq 0 \; \text{and} \; v_k = a_{jk} - d_k \geq 0.
\end{aligned}
\end{equation}

\noindent Then 

\begin{equation} \label{eq:26}
\begin{aligned}
\sum_{k} u_k = \sum_{k} v_k = 1 - e \; \text{and} \; u_{k}v_{k} = 0 \; \text{for any} \; k.
\end{aligned}
\end{equation}

\noindent Two rows are decomposed into their common and residual parts:

\begin{equation} \label{eq:27}
\begin{aligned}
\sum_{k} a_{ik} \textbf{b}_k 
= \underbrace{\sum_{k} d_k \textbf{b}_k}_{\text{common}} 
+ \sum_{k} u_k \textbf{b}_k,
\end{aligned}
\end{equation}

\begin{equation} \label{eq:28}
\begin{aligned}
\sum_{k} a_{jk} \textbf{b}_k 
= \underbrace{\sum_{k} d_k \textbf{b}_k}_{\text{common}} 
+ \sum_{k} v_k \textbf{b}_k.
\end{aligned}
\end{equation}

\noindent Subtracting \ref{eq:28} from \ref{eq:27} gives the following 

\begin{equation} \label{eq:29}
\begin{aligned}
\sum_{k} a_{ik} \textbf{b}_k - \sum_{k} a_{jk} \textbf{b}_k 
= \sum_{k} u_k \textbf{b}_k - \sum_{k} v_k \textbf{b}_k.
\end{aligned}
\end{equation}

\noindent If $e = 1$, then $u_k = v_k = 0$. \ref{eq:3} holds trivially. Consider $e \geq 1$ and then normalizing $u$, $v$:

\begin{equation} \label{eq:30}
\begin{aligned}
\tilde{u}_k = \frac{u_k}{1 - e}, \qquad
\tilde{v}_k = \frac{v_k}{1 - e}, \qquad
\sum_{k} \tilde{u}_k = \sum_{k} \tilde{v}_k = 1.
\end{aligned}
\end{equation}

\begin{equation} \label{eq:31}
\begin{aligned}
\sum_{k} u_k \textbf{b}_k - \sum_{k} v_k \textbf{b}_k 
= (1 - e) \left( \sum_{k} \tilde{u}_k \textbf{b}_k - \sum_{k} \tilde{v}_k \textbf{b}_k \right).
\end{aligned}
\end{equation}

\noindent Based on \ref{eq:25}, \ref{eq:31} is rewritten as

\begin{equation} \label{eq:32}
\begin{aligned}
\sum_{k} a_{ik} \textbf{b}_k - \sum_{k} a_{jk} \textbf{b}_k 
= (1 - e) \left( \sum_{k} \tilde{u}_k \textbf{b}_k - \sum_{k} \tilde{v}_k \textbf{b}_k \right).
\end{aligned}
\end{equation}

\noindent Applying the norm of both sides of \ref{eq:32} yields 

\begin{equation} \label{eq:33}
\begin{aligned}
\left\| \sum_{k} a_{ik} \textbf{b}_k - \sum_{k} a_{jk} \textbf{b}_k \right\|
= (1 - e) \left\| \sum_{k} \tilde{u}_k \textbf{b}_k - \sum_{k} \tilde{v}_k \textbf{b}_k \right\|.
\end{aligned}
\end{equation}

\noindent For any two convex combination of $\{\textbf{b}_k\}$

\begin{equation} \label{eq:34}
\begin{aligned}
\left\| \sum_{k} \alpha_k \textbf{b}_k - \sum_{k} \beta_k \textbf{b}_k \right\|
\leq \max_{x,y} \| \textbf{b}_x - \textbf{b}_y \|.
\end{aligned}
\end{equation}

\noindent Substituting $\alpha_k = \tilde{u}_k$ and $\beta_k = \tilde{v}_k$. From \ref{eq:33} and \ref{eq:34} yields

\begin{equation} \label{eq:35}
\begin{aligned}
\left\| \sum_{k} a_{ik} \textbf{b}_k - \sum_{k} a_{jk} \textbf{b}_k \right\|
\leq (1 - e) \max_{x,y} \| \textbf{b}_x - \textbf{b}_y \|.
\end{aligned}
\end{equation}

\noindent The inequality \ref{eq:35} is equivalent to

\begin{equation} \label{eq:36}
\begin{aligned}
\left\| \sum_{k} a_{ik} \textbf{b}_k - \sum_{k} a_{jk} \textbf{b}_k \right\|
\leq \left( 1 - \min_{i,j} \sum_{k} \min \{ a_{ik}, a_{jk} \} \right)
\max_{x,y} \| \textbf{b}_x - \textbf{b}_y \|
= \delta(\textbf{A}) \max_{x,y} \| \textbf{b}_x - \textbf{b}_y \|.
\end{aligned}
\end{equation}

\noindent Therefore, Lemma 1 has been proved.

\section{Hidden layer analysis}
\label{Appendix:C}
\begin{figure}[H]
    \centering
    \begin{minipage}[b]{0.7\textwidth}
        \centering
        \includegraphics[width=\textwidth]{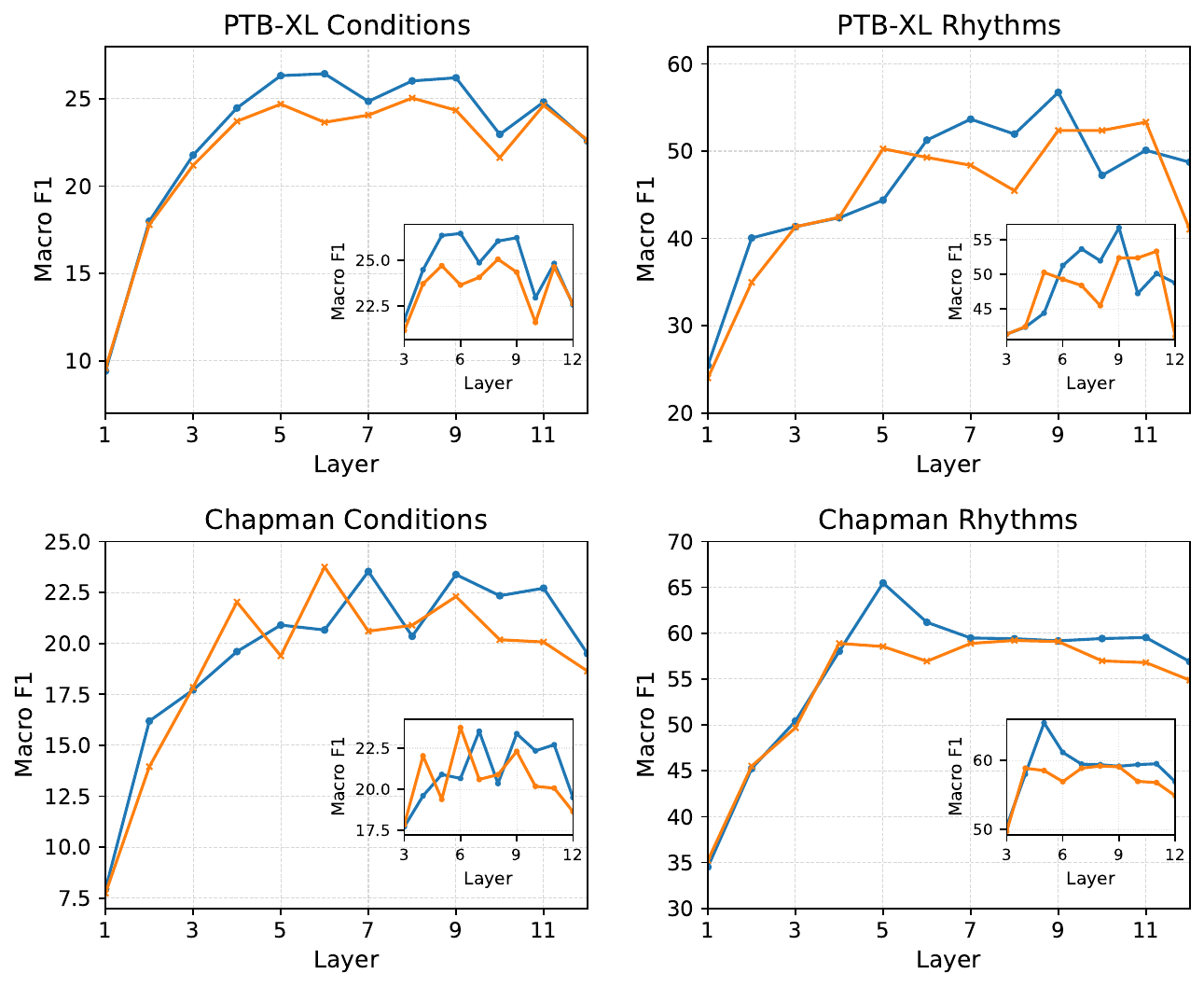}
        {c) Macro F1 score}
    \end{minipage}
    \caption{Layer-wise Performance of STMEM vs IPASTMEM on PTB-XL Conditions, PTB-XL Rhythms, Chapman Conditions, and Chapman Rhythms. We train the classification head on top of representations from pretrained ViT models for ECG condition and rhythm classification.}
    \label{fig:8}

\end{figure}

\noindent Figure \ref{fig:8} illustrates the classification performance of the STMEM and IPASTMEM models on each representation layer (from layer 1 to layer 12) on four datasets: PTB-XL Conditions, PTB-XL Rhythms, Chapman Conditions, and Chapman Rhythms. The plots show the macro AUC, instance precision, and macro F1 score values when training the classifier on the representations extracted across layers in the pretrained ViT model. The results indicate that the representations at the final layers of the model do not provide the best performance for the downstream classification tasks. The performance improves gradually from the early layers, peaks at the middle layers, and degrades at deeper layers. This trend is consistently observed across multiple evaluation metrics, datasets, and pre-trained models (i.e., STMEM and IPASTMEM), highlighting the generality of the phenomenon. Thus, we can conclude that the middle layers of the Transformer provide richer representations.

\begin{figure}[H]
    \centering
    \begin{minipage}[b]{0.7\textwidth}
        \centering
        \includegraphics[width=\textwidth]{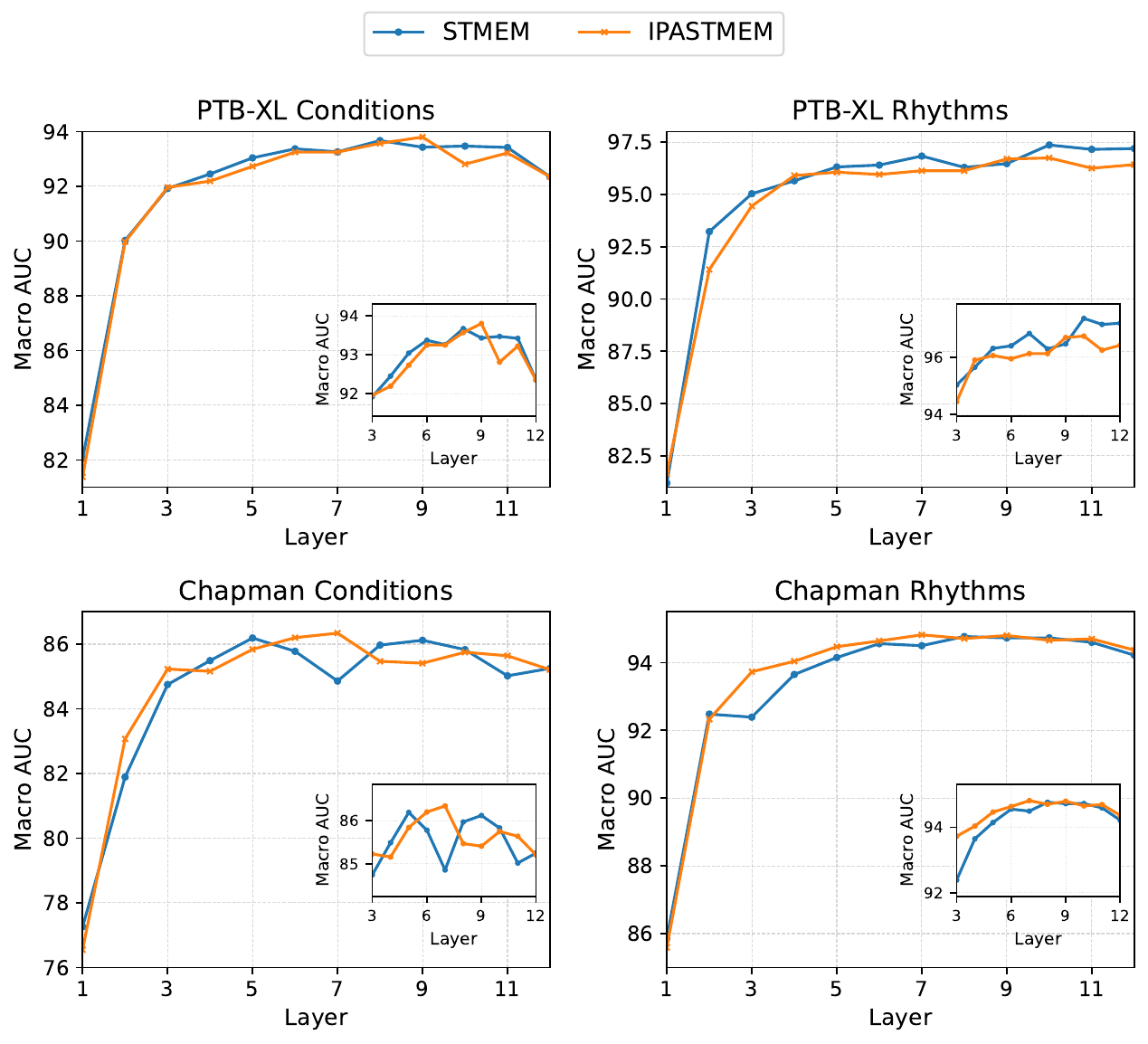}
        {a) Macro AUC}
    \end{minipage}
\end{figure}

\begin{figure}[H]
    \centering
    \begin{minipage}[b]{0.7\textwidth}
        \centering
        \includegraphics[width=\textwidth]{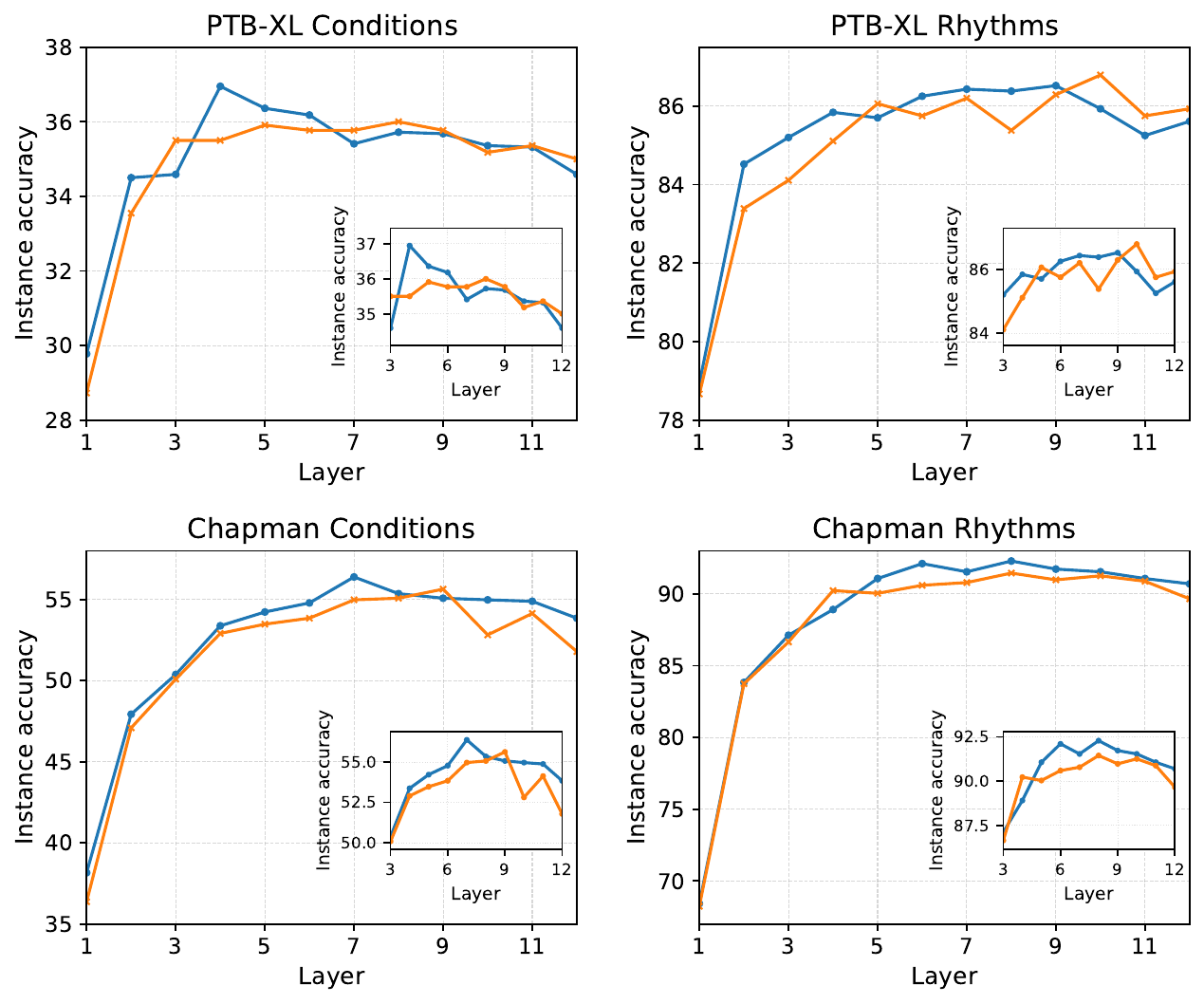}
        {b) Instance accuracy}
    \end{minipage}
\end{figure}

\section{Out-of-distribution}
\label{Appendix:D}

\renewcommand{\arraystretch}{1.4}
\begin{table}[H]
\centering
\caption{PTB-XL results under linear probing on condition and rhythm classification (out-of-distribution evaluation)}
\label{tab:ptb_out_dis}
\vspace{-0.5em} 
\scriptsize
\resizebox{0.9\textwidth}{!}{
\begin{tabular}{|l|l|l|c|c|c|c|c|c|}
\hline
\textbf{Downstream task} & \textbf{Method} & \textbf{Model} & \multicolumn{6}{c|}{\textbf{Metrics}} \\
\hline
 & & & \textbf{Macro AUC} & \textbf{Sample AUC} & \textbf{Instance Accuracy} & \textbf{Sample Accuracy} & \textbf{Macro F1 Score} & \textbf{Sample F1 Score} \\
\hline
\multirow{9}{*}{Condition}
&\multirow{3}{*}{Supervised}
 & XResnet1d50 \cite{nils2021, temesgen2022} & 90.46$\pm$0.38 & 96.36$\pm$0.13 & 34.96$\pm$1.13 & 97.79$\pm$0.03 & 21.83$\pm$0.86 & 68.94$\pm$0.65  \\
 & & ViT1d \cite{alexey2021, yeongyeon2024} & 85.31$\pm$0.45 & 94.79$\pm$0.24 & 31.27$\pm$1.9 & 97.45$\pm$0.05 & 17.47$\pm$1.6 & 63.79$\pm$1.03  \\
 & & 4FC+2LSTM+2FC \cite{temesgen2022} & 91.14$\pm$0.48 & 96.59$\pm$0.25 & 35.62$\pm$0.93 & 97.8$\pm$0.06 & 23.9$\pm$1.32 & 69.65$\pm$0.8  \\
\cline{2-9}
&\multirow{3}{*}{Self-supervised} 
 & SimCLR \cite{ting2020, temesgen2022} & 84.65$\pm$0.19 & 94.69$\pm$0.22 & 28.89$\pm$1.13 & 97.47$\pm$0.01 & 13.96$\pm$0.69 & 62.34$\pm$0.31 \\
 & & CPC \cite{temesgen2022} & 89.7$\pm$0.44 & 95.91$\pm$0.15 & 31.93$\pm$0.25 & 97.62$\pm$0.01 & 15.88$\pm$0.94 & 65.3$\pm$0.39 \\
 & & STMEM \cite{yeongyeon2024} & 92.17$\pm$0.18 & 96.89$\pm$0.12 & 34.18$\pm$0.36 & 97.79$\pm$0.01 & 21.68$\pm$1.32 & 68.14$\pm$0.31  \\
\cline{2-9}
&\multirow{3}{*}{Proposed method}
 & Scheme I - PPA & 92.73$\pm$0.34 & 97.21$\pm$0.08 & 35.26$\pm$0.45 & 97.88$\pm$0.02 & 24.77$\pm$0.71 & 69.64$\pm$0.32 \\
 & & Scheme II - PMA & 92.77$\pm$0.27 & \textbf{97.25$\pm$0.05} & 35.16$\pm$0.43 & 97.87$\pm$0.01 & \textbf{24.83$\pm$0.9} & 69.72$\pm$0.25 \\
 & & Scheme III - IPASTMEM & \textbf{93.08$\pm$0.24} & 97.2$\pm$0.09 & \textbf{35.43$\pm$0.37} & \textbf{97.89$\pm$0.01} & 24.32$\pm$1.16 & \textbf{69.94$\pm$0.18} \\
\hline
\multirow{9}{*}{Rhythm}
&\multirow{3}{*}{Supervised}
 & XResnet1d50 \cite{nils2021, temesgen2022} & 90.13$\pm$3.12 & 94.75$\pm$0.39 & 84.47$\pm$0.46 & 98.1$\pm$0.06 & 40.9$\pm$2.7 & 84.68$\pm$0.54 \\
 & & ViT1d \cite{alexey2021, yeongyeon2024} & 88.36$\pm$0.81 & 93.17$\pm$0.66 & 78.86$\pm$0.95 & 97.28$\pm$0.09 & 32.59$\pm$3.39 & 78.9$\pm$1.05 \\
 & & 4FC+2LSTM+2FC \cite{temesgen2022} & 93.94$\pm$3 & 91.72$\pm$7.84 & 78.6$\pm$14.07 & 97.59$\pm$1.12 & 37.4$\pm$8.93 & 78.43$\pm$14.87 \\
\cline{2-9}
&\multirow{3}{*}{Self-supervised} 
 & SimCLR \cite{ting2020, temesgen2022} & 84.98$\pm$0.71 & 96.17$\pm$0.36 & 81.38$\pm$0.6 & 97.36$\pm$0.05 & 24.42$\pm$0.87 & 83.32$\pm$0.43 \\
 & & CPC \cite{temesgen2022} & 90.42$\pm$0.81 & 94.25$\pm$0.11 & 81.28$\pm$0.4 & 97.59$\pm$0.05 & 32.21$\pm$3.04 & 81.54$\pm$0.32 \\
 & & STMEM \cite{yeongyeon2024} & 96.79$\pm$0.32 & 95.47$\pm$0.11 & 85.78$\pm$0.16 & 98.24$\pm$0.04 & 46.33$\pm$4.18 & 85.98$\pm$0.23 \\
\cline{2-9}
&\multirow{3}{*}{Proposed method}
 & Scheme I - PPA & 96.59$\pm$0.18 & 95.57$\pm$0.18 & \textbf{86.28$\pm$0.29} & \textbf{98.34$\pm$0.04} & \textbf{50.55$\pm$3.26} & \textbf{86.64$\pm$0.35} \\
 & & Scheme II - PMA & 96.54$\pm$0.23 & 95.53$\pm$0.13 & 86.22$\pm$0.19 & 98.33$\pm$0.03 & 50.5$\pm$2.78 & 86.53$\pm$0.26 \\
 & & Scheme III - IPASTMEM & \textbf{97.02$\pm$0.3} & \textbf{95.63$\pm$0.14} & 86.27$\pm$0.24 & 98.32$\pm$0.03 & 49.65$\pm$3.34 & 86.5$\pm$0.23 \\
\hline
\end{tabular}
}
\end{table}

\section{Ablation study}
\label{Appendix:E}

\renewcommand{\arraystretch}{2}
\begin{table}[H]
\centering
\caption{Performance gap between linear probing and fine-tuning on PTB-XL condition and rhythm classification}
\label{tab:ptb_gap}
\vspace{-0.5em}
\scriptsize
\resizebox{0.9\textwidth}{!}{
\begin{tabular}{|c|l|l|c|c|c|c|c|c|}
\hline
\textbf{Downstream task} & \textbf{Method} & \textbf{Model} & \multicolumn{6}{c|}{\textbf{Metrics}} \\
\hline
 & & & \textbf{Macro AUC} & \textbf{Sample AUC} & \textbf{Instance Accuracy} & \textbf{Sample Accuracy} & \textbf{Macro F1 Score} & \textbf{Sample F1 Score} \\
\hline
\multirow{6}{*}{Condition}
&\multirow{3}{*}{Self-supervised}
& SimCLR \cite{ting2020, temesgen2022} &
\makecell{84.97$\pm$0.28 \\ $\rightarrow$ 92.41$\pm$0.51 (+7.44)} &
\makecell{94.63$\pm$0.31 \\ $\rightarrow$ 96.86$\pm$0.18 (+2.23)} &
\makecell{29.37$\pm$1.47 \\ $\rightarrow$ 35.34$\pm$1.43 (+5.97)} &
\makecell{97.45$\pm$0.02 \\ $\rightarrow$ 97.84$\pm$0.06 (+0.39)} &
\makecell{12.43$\pm$0.62 \\ $\rightarrow$ 24.31$\pm$1.23 (+11.88)} &
\makecell{61.88$\pm$0.84 \\ $\rightarrow$ 70.28$\pm$0.65 (+8.40)} \\
& & CPC \cite{temesgen2022} &
\makecell{89.97$\pm$0.3 \\ $\rightarrow$ 91.24$\pm$0.36 (+1.27)} &
\makecell{96.04$\pm$0.09 \\ $\rightarrow$ 96.52$\pm$0.09 (+0.48)} &
\makecell{32.60$\pm$0.39 \\ $\rightarrow$ 34.61$\pm$0.22 (+2.01)} &
\makecell{97.66$\pm$0.01 \\ $\rightarrow$ 97.77$\pm$0.02 (+0.11)} &
\makecell{17.12$\pm$0.69 \\ $\rightarrow$ 19.62$\pm$0.69 (+2.50)} &
\makecell{66.23$\pm$0.2 \\ $\rightarrow$ 68.26$\pm$0.26 (+2.03)} \\
& & STMEM \cite{yeongyeon2024} &
\makecell{92.60$\pm$0.17 \\ $\rightarrow$ 93.69$\pm$0.28 (+1.09)} &
\makecell{96.92$\pm$0.07 \\ $\rightarrow$ 97.41$\pm$0.1 (+0.49)} &
\makecell{34.78$\pm$0.6 \\ $\rightarrow$ 38.26$\pm$1.06 (+3.48)} &
\makecell{97.82$\pm$0.01 \\ $\rightarrow$ 98.02$\pm$0.03 (+0.20)} &
\makecell{21.6$\pm$0.9 \\ $\rightarrow$ 26.21$\pm$1.43 (+4.61)} &
\makecell{68.41$\pm$0.35 \\ $\rightarrow$ 72.4$\pm$0.6 (+3.99)} \\
\cline{2-9}
&\multirow{3}{*}{Proposed method}
& Scheme I - PPA &
\makecell{93.39$\pm$0.29 \\ $\rightarrow$ 94.01$\pm$0.26 (+0.62)} &
\makecell{97.21$\pm$0.09 \\ $\rightarrow$ 97.55$\pm$0.06 (+0.34)} &
\makecell{35.42$\pm$0.41 \\ $\rightarrow$ 38.66$\pm$0.56 (+3.24)} &
\makecell{97.88$\pm$0.02 \\ $\rightarrow$ 98.03$\pm$0.02 (+0.15)} &
\makecell{24.43$\pm$1.41 \\ $\rightarrow$ 27.65$\pm$0.85 (+3.22)} &
\makecell{69.82$\pm$0.37 \\ $\rightarrow$ 72.85$\pm$0.36 (+3.03)} \\
& & Scheme II - PMA &
\makecell{93.44$\pm$0.14 \\ $\rightarrow$ 94.07$\pm$0.18 (+0.63)} &
\makecell{97.22$\pm$0.08 \\ $\rightarrow$ 97.6$\pm$0.15 (+0.38)} &
\makecell{35.57$\pm$0.34 \\ $\rightarrow$ 38.44$\pm$0.43 (+2.87)} &
\makecell{97.89$\pm$0.02 \\ $\rightarrow$ 98.03$\pm$0.02 (+0.14)} &
\makecell{24.73$\pm$0.78 \\ $\rightarrow$ 27.87$\pm$1.11 (+3.14)} &
\makecell{69.93$\pm$0.31 \\ $\rightarrow$ 72.77$\pm$0.28 (+2.84)} \\
& & Scheme III - IPASTMEM &
\makecell{93.2$\pm$0.3 \\ $\rightarrow$ 93.89$\pm$0.2 (+0.69)} &
\makecell{97.14$\pm$0.09 \\ $\rightarrow$ 97.44$\pm$0.17 (+0.30)} &
\makecell{35.74$\pm$0.42 \\ $\rightarrow$ 38.88$\pm$0.8 (+3.14)} &
\makecell{97.89$\pm$0.02 \\ $\rightarrow$ 98.03$\pm$0.04 (+0.14)} &
\makecell{23.53$\pm$1.11 \\ $\rightarrow$ 26.79$\pm$1.1 (+3.26)} &
\makecell{69.93$\pm$0.33 \\ $\rightarrow$ 72.79$\pm$0.62 (+2.86)} \\
\hline
\multirow{6}{*}{Rhythm}
&\multirow{3}{*}{Self-supervised}
& SimCLR \cite{ting2020, temesgen2022} &
\makecell{87.12$\pm$0.28 \\ $\rightarrow$ 96.27$\pm$0.6 (+9.15)} &
\makecell{96.28$\pm$0.48 \\ $\rightarrow$ 97.32$\pm$0.49 (+1.04)} &
\makecell{81.8$\pm$0.36 \\ $\rightarrow$ 89.05$\pm$0.78 (+7.25)} &
\makecell{97.43$\pm$0.04 \\ $\rightarrow$ 98.53$\pm$0.08 (+1.1)} &
\makecell{25.82$\pm$1.39 \\ $\rightarrow$ 44.88$\pm$2.48 (+19.06)} &
\makecell{83.77$\pm$0.63 \\ $\rightarrow$ 89.89$\pm$0.96 (+6.12)} \\
& & CPC \cite{temesgen2022} &
\makecell{91.51$\pm$0.37 \\ $\rightarrow$ 92.98$\pm$0.74 (+1.47)} &
\makecell{94.33$\pm$0.17 \\ $\rightarrow$ 94.58$\pm$0.14 (+0.25)} &
\makecell{82.05$\pm$0.23 \\ $\rightarrow$ 83.21$\pm$0.23 (+1.16)} &
\makecell{97.74$\pm$0.03 \\ $\rightarrow$ 97.91$\pm$0.04 (+0.17)} &
\makecell{30.69$\pm$2.58 \\ $\rightarrow$ 36.5$\pm$2.34 (+5.81)} &
\makecell{82.25$\pm$0.27 \\ $\rightarrow$ 83.29$\pm$0.23 (+1.04)} \\
& & STMEM \cite{yeongyeon2024} &
\makecell{97.25$\pm$0.21 \\ $\rightarrow$ 97.04$\pm$0.26 (-0.21)} &
\makecell{95.33$\pm$0.12 \\ $\rightarrow$ 95.51$\pm$0.17 (+0.18)} &
\makecell{85.55$\pm$0.38 \\ $\rightarrow$ 86.47$\pm$0.3 (+0.92)} &
\makecell{98.24$\pm$0.05 \\ $\rightarrow$ 98.36$\pm$0.04 (+0.12)} &
\makecell{44.1$\pm$2.73 \\ $\rightarrow$ 47.87$\pm$2.58 (+3.77)} &
\makecell{85.72$\pm$0.28 \\ $\rightarrow$ 86.57$\pm$0.35 (+0.85)} \\
\cline{2-9}
&\multirow{3}{*}{Proposed method}
& Scheme I - PPA &
\makecell{97.18$\pm$0.16 \\ $\rightarrow$ 96.88$\pm$0.2 (-0.3)} &
\makecell{95.5$\pm$0.12 \\ $\rightarrow$ 95.64$\pm$0.23 (+0.14)} &
\makecell{86.02$\pm$0.24 \\ $\rightarrow$ 86.79$\pm$0.31 (+0.77)} &
\makecell{98.3$\pm$0.03 \\ $\rightarrow$ 98.4$\pm$0.04 (+0.1)} &
\makecell{46.79$\pm$2.73 \\ $\rightarrow$ 48.51$\pm$2.45 (+1.72)} &
\makecell{86.28$\pm$0.24 \\ $\rightarrow$ 86.9$\pm$0.37 (+0.62)} \\
& & Scheme II - PMA &
\makecell{97.14$\pm$0.24 \\ $\rightarrow$ 96.74$\pm$0.41 (-0.4)} &
\makecell{95.68$\pm$0.11 \\ $\rightarrow$ 95.76$\pm$0.17 (+0.08)} &
\makecell{86.27$\pm$0.26 \\ $\rightarrow$ 86.67$\pm$0.33 (+0.4)} &
\makecell{98.32$\pm$0.04 \\ $\rightarrow$ 98.38$\pm$0.05 (+0.06)} &
\makecell{48.21$\pm$4.17 \\ $\rightarrow$ 50.9$\pm$2.73 (+2.69)} &
\makecell{86.6$\pm$0.15 \\ $\rightarrow$ 87.03$\pm$0.4 (+0.43)} \\
& & Scheme III - IPASTMEM &
\makecell{96.66$\pm$0.16 \\ $\rightarrow$ 96.36$\pm$0.45 (-0.3)} &
\makecell{95.7$\pm$0.12 \\ $\rightarrow$ 95.62$\pm$0.21 (-0.08)} &
\makecell{86.34$\pm$0.28 \\ $\rightarrow$ 86.74$\pm$0.4 (+0.4)} &
\makecell{98.31$\pm$0.03 \\ $\rightarrow$ 98.37$\pm$0.06 (+0.06)} &
\makecell{47.46$\pm$2.24 \\ $\rightarrow$ 48.34$\pm$3.49 (+0.88)} &
\makecell{86.61$\pm$0.25 \\ $\rightarrow$ 86.8$\pm$0.44 (+0.19)} \\
\hline
\end{tabular}
}
\end{table}

\renewcommand{\arraystretch}{2}
\begin{table}[H]
\centering
\caption{Performance gap between linear probing and fine-tuning on Chapman condition and rhythm classification}
\label{tab:chapman-gap}
\vspace{-0.5em}
\scriptsize
\resizebox{0.9\textwidth}{!}{
\begin{tabular}{|c|l|l|c|c|c|c|c|c|}
\hline
\textbf{Downstream task} & \textbf{Method} & \textbf{Model} & \multicolumn{6}{c|}{\textbf{Metrics}} \\
\hline
 & & & \textbf{Macro AUC} & \textbf{Sample AUC} & \textbf{Instance Accuracy} & \textbf{Sample Accuracy} & \textbf{Macro F1 Score} & \textbf{Sample F1 Score} \\
\hline
\multirow{6}{*}{Condition}
&\multirow{3}{*}{Self-supervised}
& SimCLR \cite{ting2020, temesgen2022} &
\makecell{77.37$\pm$0.35 \\ $\rightarrow$ 83.65$\pm$0.44 (+6.28)} &
\makecell{89.44$\pm$0.62 \\ $\rightarrow$ 97.7$\pm$0.47 (+8.26)} &
\makecell{36.35$\pm$0.72 \\ $\rightarrow$ 54.47$\pm$2.34 (+18.12)} &
\makecell{98.17$\pm$0.02 \\ $\rightarrow$ 98.79$\pm$0.08 (+0.62)} &
\makecell{12.52$\pm$0.95 \\ $\rightarrow$ 22.59$\pm$1.62 (+10.07)} &
\makecell{57.81$\pm$0.95 \\ $\rightarrow$ 79.38$\pm$1.09 (+21.57)} \\
& & CPC \cite{temesgen2022} &
\makecell{81.96$\pm$0.48 \\ $\rightarrow$ 82.69$\pm$0.34 (+0.73)} &
\makecell{95.19$\pm$0.26 \\ $\rightarrow$ 96.13$\pm$0.31 (+0.94)} &
\makecell{47.28$\pm$0.63 \\ $\rightarrow$ 49.85$\pm$0.61 (+2.57)} &
\makecell{98.51$\pm$0.02 \\ $\rightarrow$ 98.61$\pm$0.03 (+0.1)} &
\makecell{14.87$\pm$0.68 \\ $\rightarrow$ 16.56$\pm$0.61 (+1.69)} &
\makecell{70.4$\pm$0.47 \\ $\rightarrow$ 73.37$\pm$0.5 (+2.97)} \\
& & STMEM \cite{yeongyeon2024} &
\makecell{85.24$\pm$0.22 \\ $\rightarrow$ 85.37$\pm$0.25 (+0.13)} &
\makecell{97.76$\pm$0.24 \\ $\rightarrow$ 98.22$\pm$0.23 (+0.46)} &
\makecell{53.58$\pm$0.66 \\ $\rightarrow$ 55.42$\pm$1.02 (+1.84)} &
\makecell{98.77$\pm$0.01 \\ $\rightarrow$ 98.84$\pm$0.03 (+0.07)} &
\makecell{19.76$\pm$1.12 \\ $\rightarrow$ 21.82$\pm$0.95 (+2.06)} &
\makecell{78.5$\pm$0.52 \\ $\rightarrow$ 80.39$\pm$0.77 (+1.89)} \\
\cline{2-9}
&\multirow{3}{*}{Proposed method}
& Scheme I - PPA &
\makecell{85.63$\pm$0.28 \\ $\rightarrow$ 85.72$\pm$0.27 (+0.09)} &
\makecell{98.16$\pm$0.19 \\ $\rightarrow$ 98.25$\pm$0.19 (+0.09)} &
\makecell{55.41$\pm$0.54 \\ $\rightarrow$ 56.05$\pm$0.59 (+0.64)} &
\makecell{98.85$\pm$0.01 \\ $\rightarrow$ 98.87$\pm$0.01 (+0.02)} &
\makecell{21.45$\pm$0.45 \\ $\rightarrow$ 22.8$\pm$0.49 (+1.35)} &
\makecell{80.19$\pm$0.3 \\ $\rightarrow$ 80.95$\pm$0.46 (+0.76)} \\
& & Scheme II - PMA &
\makecell{85.69$\pm$0.3 \\ $\rightarrow$ 85.76$\pm$0.27 (-0.07)} &
\makecell{98.18$\pm$0.23 \\ $\rightarrow$ 98.16$\pm$0.3 (-0.02)} &
\makecell{55.43$\pm$0.38 \\ $\rightarrow$ 55.84$\pm$0.97 (+0.41)} &
\makecell{98.85$\pm$0.01 \\ $\rightarrow$ 98.87$\pm$0.03 (+0.02)} &
\makecell{21.29$\pm$0.52 \\ $\rightarrow$ 22.97$\pm$0.92 (+1.68)} &
\makecell{80.25$\pm$0.38 \\ $\rightarrow$ 80.65$\pm$0.84 (+0.4)} \\
& & Scheme III - IPASTMEM &
\makecell{85.72$\pm$0.3 \\ $\rightarrow$ 85.84$\pm$0.22 (+0.12)} &
\makecell{97.91$\pm$0.25 \\ $\rightarrow$ 98.02$\pm$0.37 (+0.11)} &
\makecell{54.4$\pm$0.66 \\ $\rightarrow$ 55.09$\pm$0.9 (+0.69)} &
\makecell{98.82$\pm$0.02 \\ $\rightarrow$ 98.84$\pm$0.03 (+0.02)} &
\makecell{20.72$\pm$1.3 \\ $\rightarrow$ 22.31$\pm$1.39 (+1.59)} &
\makecell{79.29$\pm$0.61 \\ $\rightarrow$ 79.97$\pm$0.87 (+0.68)} \\
\hline
\multirow{6}{*}{Rhythm}
&\multirow{3}{*}{Self-supervised}
& SimCLR \cite{ting2020, temesgen2022} &
\makecell{83.31$\pm$0.24 \\ $\rightarrow$ 93.87$\pm$0.8 (+10.56)} &
\makecell{86.64$\pm$0.59 \\ $\rightarrow$ 97.1$\pm$0.73 (+10.46)} &
\makecell{61.96$\pm$1.24 \\ $\rightarrow$ 89.35$\pm$1.28 (+27.39)} &
\makecell{95.55$\pm$0.09 \\ $\rightarrow$ 98.51$\pm$0.15 (+2.96)} &
\makecell{37.28$\pm$1.52 \\ $\rightarrow$ 55.75$\pm$2.42 (+18.47)} &
\makecell{64.88$\pm$1.07 \\ $\rightarrow$ 89.97$\pm$1.43 (+25.09)} \\
& & CPC \cite{temesgen2022} &
\makecell{91.47$\pm$0.36 \\ $\rightarrow$ 92.39$\pm$0.73 (+0.92)} &
\makecell{94.03$\pm$0.62 \\ $\rightarrow$ 95.31$\pm$0.31 (+1.28)} &
\makecell{80.18$\pm$1.25 \\ $\rightarrow$ 84.04$\pm$0.75 (+3.86)} &
\makecell{97.32$\pm$0.11 \\ $\rightarrow$ 97.78$\pm$0.13 (+0.46)} &
\makecell{45.81$\pm$0.7 \\ $\rightarrow$ 47.96$\pm$1.22 (+2.15)} &
\makecell{81.2$\pm$1.12 \\ $\rightarrow$ 84.66$\pm$0.8 (+3.46)} \\
& & STMEM \cite{yeongyeon2024} &
\makecell{94.42$\pm$0.07 \\ $\rightarrow$ 94.4$\pm$0.13 (-0.02)} &
\makecell{97.6$\pm$0.16 \\ $\rightarrow$ 97.9$\pm$0.24 (+0.3)} &
\makecell{90.44$\pm$0.47 \\ $\rightarrow$ 91.35$\pm$1.02 (+0.91)} &
\makecell{98.67$\pm$0.06 \\ $\rightarrow$ 98.75$\pm$0.14 (+0.08)} &
\makecell{57.99$\pm$0.88 \\ $\rightarrow$ 59.04$\pm$1.76 (+1.05)} &
\makecell{91.16$\pm$0.29 \\ $\rightarrow$ 91.86$\pm$0.87 (+0.7)} \\
\cline{2-9}
&\multirow{3}{*}{Proposed method}
& Scheme I - PPA &
\makecell{94.68$\pm$0.1 \\ $\rightarrow$ 94.65$\pm$0.09 (-0.03)} &
\makecell{97.85$\pm$0.22 \\ $\rightarrow$ 97.87$\pm$0.3 (+0.02)} &
\makecell{91.79$\pm$0.58 \\ $\rightarrow$ 91.38$\pm$1.13 (-0.41)} &
\makecell{98.85$\pm$0.06 \\ $\rightarrow$ 98.78$\pm$0.15 (-0.07)} &
\makecell{59.19$\pm$1.19 \\ $\rightarrow$ 60.08$\pm$0.62 (+0.89)} &
\makecell{92.23$\pm$0.39 \\ $\rightarrow$ 91.96$\pm$0.94 (-0.27)} \\
& & Scheme II - PMA &
\makecell{94.65$\pm$0.07 \\ $\rightarrow$ 94.61$\pm$0.12 (-0.04)} &
\makecell{97.88$\pm$0.22 \\ $\rightarrow$ 97.93$\pm$0.29 (+0.05)} &
\makecell{91.71$\pm$0.55 \\ $\rightarrow$ 91.72$\pm$0.68 (+0.01)} &
\makecell{98.84$\pm$0.06 \\ $\rightarrow$ 98.81$\pm$0.08 (-0.03)} &
\makecell{59.33$\pm$1.12 \\ $\rightarrow$ 60.33$\pm$2.17 (+1)} &
\makecell{92.24$\pm$0.42 \\ $\rightarrow$ 92.18$\pm$0.6 (-0.06)} \\
& & Scheme III - IPASTMEM &
\makecell{94.67$\pm$0.11 \\ $\rightarrow$ 94.57$\pm$0.1 (-0.1)} &
\makecell{97.82$\pm$0.23 \\ $\rightarrow$ 97.89$\pm$0.36 (+0.07)} &
\makecell{91.51$\pm$0.39 \\ $\rightarrow$ 91.18$\pm$0.96 (-0.33)} &
\makecell{98.8$\pm$0.04 \\ $\rightarrow$ 98.73$\pm$0.15 (-0.07)} &
\makecell{58.77$\pm$0.76 \\ $\rightarrow$ 61$\pm$3.24 (+2.23)} &
\makecell{91.89$\pm$0.45 \\ $\rightarrow$ 91.82$\pm$0.99 (-0.07)} \\
\hline
\end{tabular}
}
\end{table}

\renewcommand{\arraystretch}{2}
\begin{table}[H]
\centering
\caption{Performance gap between linear probing and fine-tuning on PTB-XL condition and rhythm classification (out-of-distribution evaluation)}
\label{tab:ptbxl-gap-ood}
\vspace{-0.5em}
\scriptsize
\resizebox{0.9\textwidth}{!}{
\begin{tabular}{|c|l|l|c|c|c|c|c|c|}
\hline
\textbf{Downstream task} & \textbf{Method} & \textbf{Model} & \multicolumn{6}{c|}{\textbf{Metrics}} \\
\hline
 & & & \textbf{Macro AUC} & \textbf{Sample AUC} & \textbf{Instance Accuracy} & \textbf{Sample Accuracy} & \textbf{Macro F1 Score} & \textbf{Sample F1 Score} \\
\hline
\multirow{6}{*}{Condition}
&\multirow{3}{*}{Self-supervised}
& SimCLR \cite{ting2020, temesgen2022} &
\makecell{84.65$\pm$0.19 \\ $\rightarrow$ 92.38$\pm$0.42 (+7.73)} &
\makecell{94.69$\pm$0.22 \\ $\rightarrow$ 96.96$\pm$0.19 (+2.27)} &
\makecell{28.89$\pm$1.13 \\ $\rightarrow$ 34.75$\pm$1.51 (+5.86)} &
\makecell{97.47$\pm$0.01 \\ $\rightarrow$ 97.85$\pm$0.04 (+0.38)} &
\makecell{13.96$\pm$0.69 \\ $\rightarrow$ 24.51$\pm$2.13 (+10.55)} &
\makecell{62.34$\pm$0.31 \\ $\rightarrow$ 70.43$\pm$0.73 (+8.09)} \\
& & CPC \cite{temesgen2022} &
\makecell{89.7$\pm$0.44 \\ $\rightarrow$ 90.64$\pm$0.27 (+0.94)} &
\makecell{95.91$\pm$0.15 \\ $\rightarrow$ 96.35$\pm$0.07 (+0.44)} &
\makecell{31.93$\pm$0.25 \\ $\rightarrow$ 33.8$\pm$0.31 (+1.87)} &
\makecell{97.62$\pm$0.01 \\ $\rightarrow$ 97.74$\pm$0.01 (+0.12)} &
\makecell{15.88$\pm$0.94 \\ $\rightarrow$ 19.33$\pm$0.39 (+3.45)} &
\makecell{65.3$\pm$0.39 \\ $\rightarrow$ 67.75$\pm$0.22 (+2.45)} \\
& & STMEM \cite{yeongyeon2024} &
\makecell{92.17$\pm$0.18 \\ $\rightarrow$ 93.56$\pm$0.29 (+1.39)} &
\makecell{96.89$\pm$0.12 \\ $\rightarrow$ 97.43$\pm$0.09 (+0.54)} &
\makecell{34.18$\pm$0.36 \\ $\rightarrow$ 38.23$\pm$0.75 (+4.05)} &
\makecell{97.79$\pm$0.01 \\ $\rightarrow$ 98.02$\pm$0.01 (+0.23)} &
\makecell{21.68$\pm$1.32 \\ $\rightarrow$ 27.01$\pm$1.06 (+5.33)} &
\makecell{68.14$\pm$0.31 \\ $\rightarrow$ 72.57$\pm$0.43 (+4.43)} \\
\cline{2-9}
&\multirow{3}{*}{Proposed method}
& Scheme I - PPA &
\makecell{92.73$\pm$0.34 \\ $\rightarrow$ 93.82$\pm$0.22 (+1.09)} &
\makecell{97.21$\pm$0.08 \\ $\rightarrow$ 97.48$\pm$0.12 (+0.27)} &
\makecell{35.26$\pm$0.45 \\ $\rightarrow$ 38.22$\pm$0.81 (+2.96)} &
\makecell{97.88$\pm$0.02 \\ $\rightarrow$ 98.02$\pm$0.02 (+0.14)} &
\makecell{24.77$\pm$0.71 \\ $\rightarrow$ 27.97$\pm$0.55 (+3.2)} &
\makecell{69.64$\pm$0.32 \\ $\rightarrow$ 72.6$\pm$0.39 (+2.96)} \\
& & Scheme II - PMA &
\makecell{92.77$\pm$0.27 \\ $\rightarrow$ 93.92$\pm$0.12 (+1.15)} &
\makecell{97.25$\pm$0.05 \\ $\rightarrow$ 97.47$\pm$0.12 (+0.22)} &
\makecell{35.16$\pm$0.43 \\ $\rightarrow$ 38.38$\pm$0.59 (+3.22)} &
\makecell{97.87$\pm$0.01 \\ $\rightarrow$ 98.01$\pm$0.03 (+0.14)} &
\makecell{24.83$\pm$0.9 \\ $\rightarrow$ 28.31$\pm$0.97 (+3.48)} &
\makecell{69.72$\pm$0.25 \\ $\rightarrow$ 72.52$\pm$0.4 (+2.80)} \\
& & Scheme III - IPASTMEM &
\makecell{93.08$\pm$0.24 \\ $\rightarrow$ 94.04$\pm$0.29 (+0.96)} &
\makecell{97.2$\pm$0.09 \\ $\rightarrow$ 97.54$\pm$0.05 (+0.34)} &
\makecell{35.43$\pm$0.37 \\ $\rightarrow$ 38.9$\pm$0.63 (+3.47)} &
\makecell{97.89$\pm$0.01 \\ $\rightarrow$ 98.04$\pm$0.02 (+0.15)} &
\makecell{24.32$\pm$1.16 \\ $\rightarrow$ 27.79$\pm$1.44 (+3.47)} &
\makecell{69.94$\pm$0.18 \\ $\rightarrow$ 73.08$\pm$0.44 (+3.14)} \\
\hline
\multirow{6}{*}{Rhythm}
&\multirow{3}{*}{Self-supervised}
& SimCLR \cite{ting2020, temesgen2022} &
\makecell{84.98$\pm$0.71 \\ $\rightarrow$ 96.44$\pm$0.45 (+11.46)} &
\makecell{96.17$\pm$0.36 \\ $\rightarrow$ 97.41$\pm$0.43 (+1.24)} &
\makecell{81.38$\pm$0.6 \\ $\rightarrow$ 89.09$\pm$0.73 (+7.71)} &
\makecell{97.36$\pm$0.05 \\ $\rightarrow$ 98.54$\pm$0.09 (+1.18)} &
\makecell{24.42$\pm$0.87 \\ $\rightarrow$ 44.28$\pm$2.87 (+19.86)} &
\makecell{83.32$\pm$0.43 \\ $\rightarrow$ 90.06$\pm$0.82 (+6.74)} \\
& & CPC \cite{temesgen2022} &
\makecell{90.42$\pm$0.81 \\ $\rightarrow$ 92.11$\pm$1.37 (+1.69)} &
\makecell{94.25$\pm$0.11 \\ $\rightarrow$ 94.4$\pm$0.07 (+0.15)} &
\makecell{81.28$\pm$0.4 \\ $\rightarrow$ 82.22$\pm$0.24 (+0.94)} &
\makecell{97.59$\pm$0.05 \\ $\rightarrow$ 97.73$\pm$0.05 (+0.14)} &
\makecell{32.21$\pm$3.04 \\ $\rightarrow$ 33.89$\pm$2.37 (+1.68)} &
\makecell{81.54$\pm$0.32 \\ $\rightarrow$ 82.37$\pm$0.16 (+0.83)} \\
& & STMEM \cite{yeongyeon2024} &
\makecell{96.79$\pm$0.32 \\ $\rightarrow$ 96.83$\pm$0.28 (+0.04)} &
\makecell{95.47$\pm$0.11 \\ $\rightarrow$ 95.51$\pm$0.2 (+0.04)} &
\makecell{85.78$\pm$0.16 \\ $\rightarrow$ 86.49$\pm$0.35 (+0.71)} &
\makecell{98.24$\pm$0.04 \\ $\rightarrow$ 98.38$\pm$0.03 (+0.14)} &
\makecell{46.33$\pm$4.18 \\ $\rightarrow$ 49.44$\pm$3.26 (+3.11)} &
\makecell{85.98$\pm$0.23 \\ $\rightarrow$ 86.69$\pm$0.35 (+0.71)} \\
\cline{2-9}
&\multirow{3}{*}{Proposed method}
& Scheme I - PPA &
\makecell{96.59$\pm$0.18 \\ $\rightarrow$ 96.55$\pm$0.3 (-0.04)} &
\makecell{95.57$\pm$0.18 \\ $\rightarrow$ 95.59$\pm$0.26 (+0.02)} &
\makecell{86.28$\pm$0.29 \\ $\rightarrow$ 86.51$\pm$0.34 (+0.23)} &
\makecell{98.34$\pm$0.04 \\ $\rightarrow$ 98.38$\pm$0.05 (+0.04)} &
\makecell{50.55$\pm$3.26 \\ $\rightarrow$ 50.43$\pm$2.89 (-0.12)} &
\makecell{86.64$\pm$0.35 \\ $\rightarrow$ 86.79$\pm$0.35 (+0.15)} \\
& & Scheme II - PMA &
\makecell{96.54$\pm$0.23 \\ $\rightarrow$ 96.08$\pm$0.38 (-0.46)} &
\makecell{95.53$\pm$0.13 \\ $\rightarrow$ 95.72$\pm$0.18 (+0.19)} &
\makecell{86.22$\pm$0.19 \\ $\rightarrow$ 86.85$\pm$0.27 (+0.63)} &
\makecell{98.33$\pm$0.03 \\ $\rightarrow$ 98.4$\pm$0.03 (+0.07)} &
\makecell{50.5$\pm$2.78 \\ $\rightarrow$ 49.01$\pm$2.98 (-1.49)} &
\makecell{86.53$\pm$0.26 \\ $\rightarrow$ 87.1$\pm$0.28 (+0.57)} \\
& & Scheme III - IPASTMEM &
\makecell{97.02$\pm$0.3 \\ $\rightarrow$ 96.73$\pm$0.42 (-0.29)} &
\makecell{95.63$\pm$0.14 \\ $\rightarrow$ 95.56$\pm$0.23 (-0.07)} &
\makecell{86.27$\pm$0.24 \\ $\rightarrow$ 86.36$\pm$0.45 (+0.09)} &
\makecell{98.32$\pm$0.03 \\ $\rightarrow$ 98.35$\pm$0.05 (+0.03)} &
\makecell{49.65$\pm$3.34 \\ $\rightarrow$ 49.92$\pm$3.19 (+0.27)} &
\makecell{86.5$\pm$0.23 \\ $\rightarrow$ 86.58$\pm$0.36 (+0.08)} \\
\hline
\end{tabular}
}
\end{table}
\end{document}